%% file: main.tex
\pdfoutput=1
\PassOptionsToPackage{table}{xcolor}
\documentclass[logo]{sofa_lab_2026}

\graphicspath{{./}}

\usepackage[numbers,sort&compress]{natbib}
\usepackage{bm}
\usepackage{nicefrac}
\input{math_commands.tex}

\usepackage{float}
\usepackage{wrapfig}
\usepackage{subfig}
\usepackage{tikz}
\usetikzlibrary{arrows.meta,positioning,fit,backgrounds,calc,shapes.geometric}
\usepackage{pgfplots}
\pgfplotsset{compat=1.18}

\usepackage{siunitx}
\usepackage{multirow}
\usepackage{adjustbox}
\usepackage{arydshln}
\usepackage[normalem]{ulem}
\useunder{\uline}{\ul}{}
\usepackage{xspace}
\usepackage{xparse}

\usepackage{listings}
\usepackage{algorithm}
\usepackage{algpseudocode}

\definecolor{ink}{HTML}{1A1D23}
\definecolor{ink2}{HTML}{5B616B}
\colorlet{clay}{BrandAccent}
\colorlet{claydark}{BrandPrimary}
\definecolor{sky}{HTML}{0A509E}
\definecolor{sage}{HTML}{4F7F3A}
\definecolor{heather}{HTML}{7A5AA6}
\colorlet{kraft}{BrandAccent}
\definecolor{ivory}{HTML}{F4F3F0}
\definecolor{ivoryline}{HTML}{D8D5CE}
\definecolor{oursbg}{HTML}{E4EBF3}
\colorlet{navy}{BrandLink}
\definecolor{pend}{HTML}{8A8F98}
\colorlet{rule}{BrandRule}
\definecolor{backcolour}{rgb}{0.95,0.95,0.92}
\definecolor{codegreen}{rgb}{0,0.6,0}
\definecolor{codegray}{rgb}{0.5,0.5,0.5}
\definecolor{codepurple}{rgb}{0.58,0,0.82}
\definecolor{methodblue}{HTML}{2457A6}
\definecolor{takeawayblue}{HTML}{2B5F97}
\definecolor{mnDarkGray}{HTML}{4D4D4D}
\definecolor{tableInk}{HTML}{263B4A}
\definecolor{tableRule}{HTML}{74828B}
\definecolor{tableHeader}{HTML}{DCE7ED}
\definecolor{tableSubhead}{HTML}{EDF3F6}
\definecolor{tableGroup}{HTML}{F2F5F6}
\definecolor{tableBand}{HTML}{F7F9FA}
\definecolor{tableOurs}{HTML}{E2EDF3}
\definecolor{takeawayink}{HTML}{294861}
\definecolor{takeawayline}{HTML}{8FA9BC}
\definecolor{takeawayback}{HTML}{E8E6E1}
\definecolor{takeawaytitleback}{HTML}{D8D5CE}

\arrayrulecolor{rule}
\newcommand{\tsz}{\footnotesize}

\newcommand{\oursrow}{\rowcolor{oursbg}}

\newcommand{\method}{\textsc{HarnessTTT}\xspace}

\newcommand{\sftm}{\operatorname{softmax}}

\algrenewcommand\algorithmicrequire{\textbf{Input:}}
\algrenewcommand\algorithmicensure{\textbf{Output:}}
\algrenewcommand\algorithmiccomment[1]{\hfill\textcolor{gray}{$\triangleright$ \footnotesize #1}}

\algnewcommand{\LineComment}[1]{\State \textcolor{gray}{$\triangleright$ \footnotesize #1}}

\tcbset{
  takeawaybox/.style={
    width=\linewidth, enhanced, breakable,
    rounded corners, arc=2.8mm, outer arc=2.8mm,
    boxrule=0.7pt,
    top=5pt, bottom=5pt, left=7pt, right=7pt,
    before skip=7pt, after skip=7pt,
    colback=takeawayback, colframe=takeawayline, coltext=black,
    colbacktitle=takeawaytitleback, coltitle=black,
    fontupper=\itshape, fonttitle=\bfseries\itshape\small,
    titlerule=0.45pt, toptitle=4pt, bottomtitle=4pt,
    lefttitle=7pt, righttitle=7pt,
  }
}
\newtcolorbox{takeawaybox}[2][]{takeawaybox, title=#2, #1}
\definecolor{groupbg}{RGB}{232,232,232} 
\ExplSyntaxOn
\NewDocumentCommand \hname { m }
  {
    \group_begin:
      \tl_set:Nn \l_tmpa_tl {#1}
      \tl_replace_all:Nnn \l_tmpa_tl { - } { - \allowbreak }
      \cs_set:Npn \_ { \textunderscore \allowbreak }
      \ttfamily
      \tl_use:N \l_tmpa_tl
    \group_end:
  }
\ExplSyntaxOff
\newcommand{\ctool}[1]{\textcolor{sky}{\hname{#1}}}
\newcommand{\cskill}[1]{\textcolor{heather}{\hname{#1}}}
\newcommand{\cprompt}[1][prompt only]{\textit{#1}}
\newcommand{\lgy}{\textcolor{sage}{\ding{51}}}
\newcommand{\lgn}{\textcolor{ink2}{\ding{55}}}

\lstdefinestyle{harnessbase}{
  basicstyle=\ttfamily\scriptsize\color{ink},
  backgroundcolor=\color{ivory},
  frame=l, framerule=1.6pt, framesep=6pt, rulecolor=\color{ivoryline},
  xleftmargin=7.6pt, xrightmargin=0pt,
  aboveskip=9pt, belowskip=7pt,
  captionpos=t, abovecaptionskip=2pt, belowcaptionskip=5pt,
  columns=fullflexible, keepspaces=true,
  breaklines=true, breakatwhitespace=false,
  postbreak=\mbox{\textcolor{ink2}{$\hookrightarrow$}\space},
  showstringspaces=false, upquote=true, tabsize=2,
}
\lstdefinestyle{harness}{style=harnessbase,
  rulecolor=\color{BrandPrimary},
  moredelim=[l][\color{BrandPrimary}\bfseries]{\#},
  moredelim=[s][\bfseries]{**}{**},
}
\lstdefinestyle{harnessyaml}{style=harnessbase,
  rulecolor=\color{kraft},
    morestring=[b]", stringstyle=\color{sage!80!black},
  morekeywords={true,false,null}, keywordstyle=\color{navy},
}
\lstdefinestyle{harnesspy}{style=harnessbase,
  language=Python,
  rulecolor=\color{BrandTint},
  keywordstyle=\color{navy}\bfseries,
  morekeywords={True,False,None},
  commentstyle=\color{ink2},
  stringstyle=\color{sage!80!black},
}

\title{\method: Learn to Optimize Harness at Test-Time for Scientific Discovery}

\author[1]{Yingzi Ma}
\author[2]{Yulong Cao}
\author[3]{Nanxi Li}
\author[4]{Fang Wu}
\author[5]{Yizhe Zhang}
\author[4]{James Zou}
\author[4]{Yejin Choi}
\author[3]{Chaowei Xiao}
\affil[1]{University of Wisconsin-Madison}
\affil[2]{NVIDIA}
\affil[3]{Johns Hopkins University}
\affil[4]{Stanford University}
\affil[5]{Meta}

\newcommand{\badgegap}{\hspace{0.6cm}}
\newcommand{\badge}[3]{%
  \href{#2}{%
    \IfFileExists{#1}%
      {\raisebox{-0.2\height}{\includegraphics[height=1em]{#1}}\hspace{0.3em}}%
      {}%
    \texttt{#3}%
  }%
}

\newcommand{\projectbadges}{%
  \badge{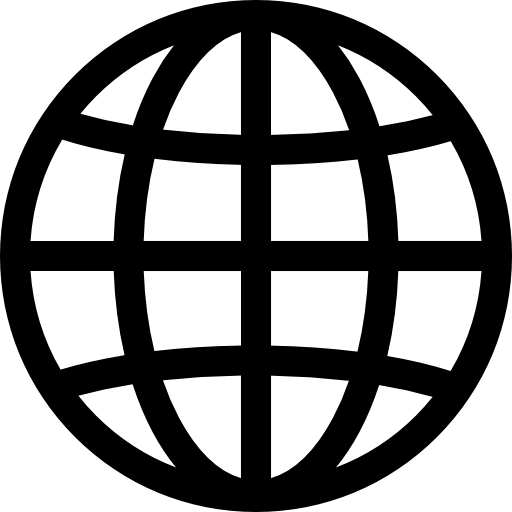}{https://gray311.github.io/harnessrl/}{Website}\badgegap
  \badge{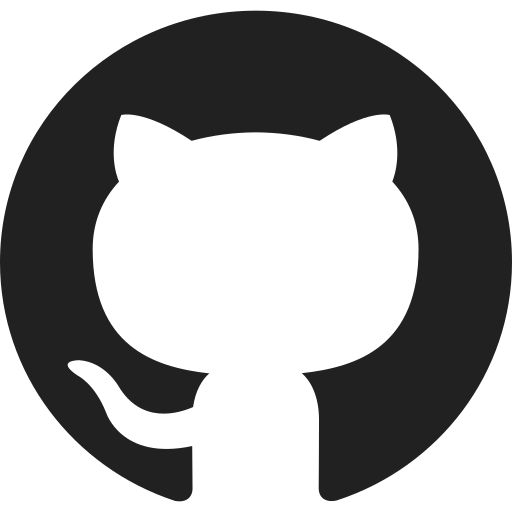}{https://github.com/SaFo-Lab/Self_Adapt_Harness}{Code}%
}

\begin{document}
\maketitle

\begin{abstract}
Scientific discovery with language models requires finding the best valid solution within a finite computational budget. We study how feedback from this search can improve the procedure used to conduct it.
We introduce \method{}, a test-time learning framework that trains a harness proposer while keeping the executor frozen. The proposer generates executable harnesses comprising system prompts, reusable skills, tools, middleware, and execution controls. Each harness guides a budgeted discovery rollout, whose outcome provides feedback for updating the proposer. A discovery-oriented policy-gradient objective combines gap-normalized rewards, leave-one-out advantages (RLOO), and max-sharpening to favor high-performing proposals. Generated components are validated before execution, and the search retains the best valid solution and selected harnesses across rounds. Across fourteen discovery tasks in mathematics,
systems, GPU kernels, heuristic programming, astrodynamics, and biology, \method{} with a
frozen 9B executor improves on OpenEvolve with the same executor on all
eleven benchmark tasks (e.g., from 1.17 to 2.64 on circle packing), improves
on the discovery-fine-tuned Finch-9B by up to $92\%$, holds the best
open-weight sub-10B result on nine, and reaches $0.380919$ on Erd\H{o}s
minimum overlap, better than the best human reference of $0.380927$.
Scaling the proposer and frozen executor to 27B improves on eleven of fourteen tasks,
and transferring the learned harnesses to GPT-5.5 matches or exceeds
the strongest reported system of any backbone on nine of fourteen tasks.
Together, these results show that harness
search can itself be learned at test time: a trained proposer lets a small
frozen executor search effectively, and the harnesses it finds carry over to
stronger executors.
\end{abstract}

\noindent\begin{minipage}{\textwidth}
\centering
\includegraphics[width=\textwidth]{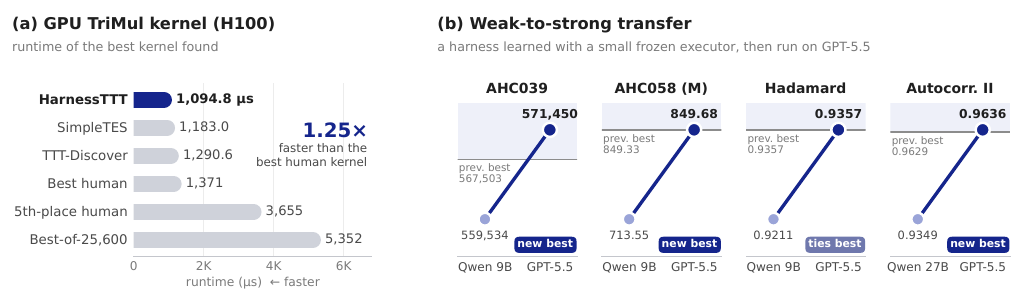}
\captionof{figure}{\textbf{\method{} at a glance.}
\textbf{(a)}~Runtime of the best TriMul kernel on an H100; \method{} runs its 27B-learned harness on GPT-5.5.
SimpleTES~\citep{ye2026simpletes} and TTT-Discover~\citep{tttdiscover} are re-measured by us; the human
and best-of-25{,}600 baselines are as reported by TTT-Discover.
\textbf{(b)}~Weak-to-strong transfer on four tasks of Table~\ref{tab:frontier-transfer}: a harness
learned with a frozen Qwen executor (9B or 27B) continues its search on GPT-5.5.}
\label{fig:teaser}
\end{minipage}

\input{subtex/intro}
\input{subtex/related}
\input{subtex/method}
\input{subtex/experiment}
\input{subtex/conclusion}

\bibliography{references}
\bibliographystyle{plainnat}

\clearpage
\appendix
\input{subtex/appendix}

\end{document}

%% file: math_commands.tex
\usepackage{amsmath,amsfonts,bm}

\def\eqref#1{equation~\ref{#1}}

\def\1{\bm{1}}

\DeclareMathAlphabet{\mathsfit}{\encodingdefault}{\sfdefault}{m}{sl}
\SetMathAlphabet{\mathsfit}{bold}{\encodingdefault}{\sfdefault}{bx}{n}

%% file: subtex/intro.tex
\begin{figure}[!t]
\centering
\includegraphics[width=\textwidth,height=0.88\textheight,keepaspectratio]{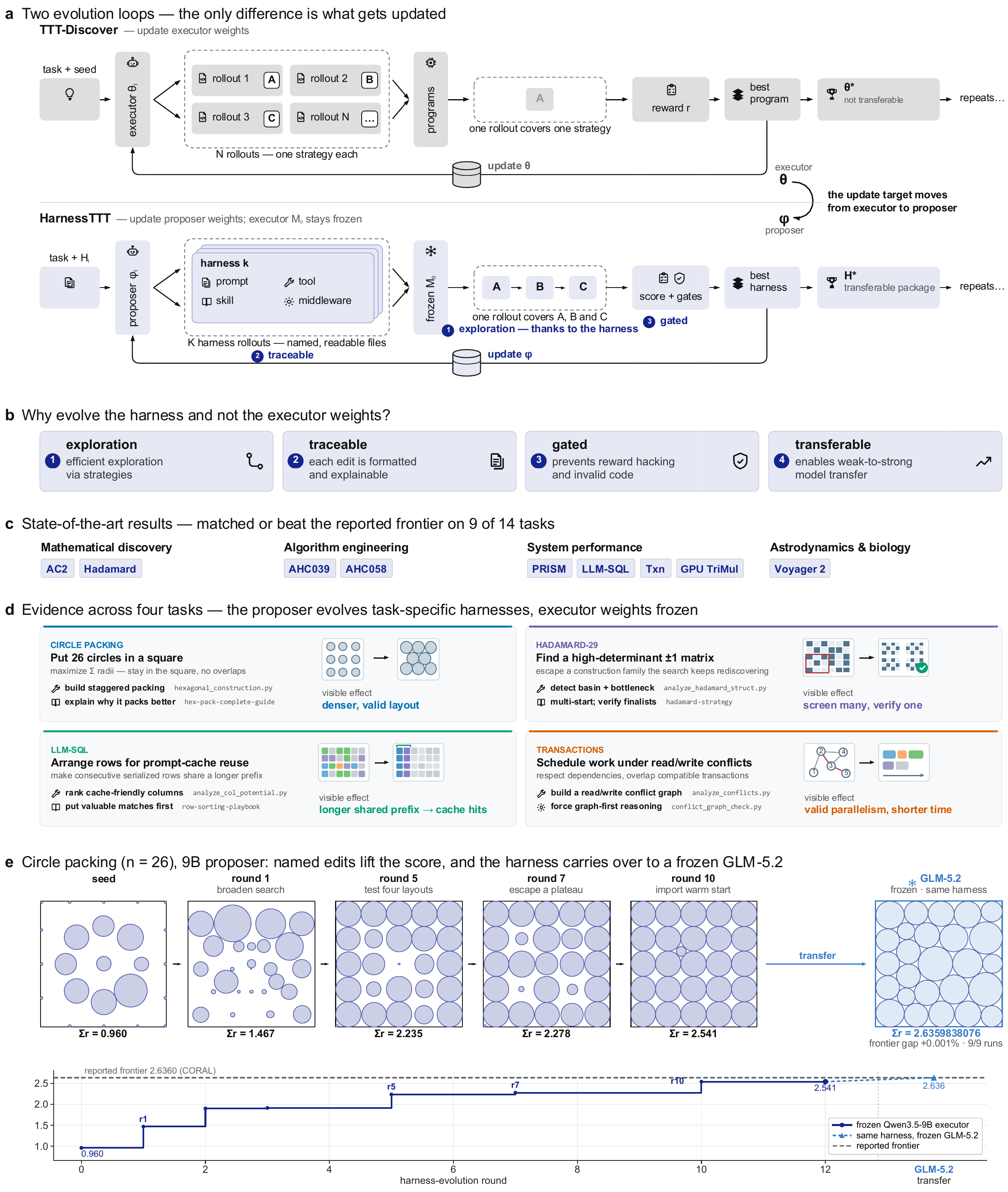}
\caption{\textbf{\method{} on one page.}
\textbf{(a)}~Executor RL versus \method{}: the same loop, but the reward
updates the executor or the proposer.
\textbf{(b)}~Properties of a learned harness.
\textbf{(c)}~Tasks at or above the reported frontier.
\textbf{(d)}~Task-specific harnesses written by the proposer.
\textbf{(e)}~A circle-packing campaign and its transfer to a frozen GLM-5.2
executor.}
\label{fig:one}
\end{figure}
\clearpage

\section{Introduction}

Scientific discovery advances knowledge and technology by finding solutions beyond what established methods can achieve~\citep{funsearch,alphaevolve}. Recent work~\citep{tttdiscover,ye2026simpletes,thetaevolve} on AI for scientific discovery formulates each task as an exploration for the best valid solution to a single challenging scientific problem within a finite computational budget, with the hope of surpassing the best-known results from the best human expert~\citep{alphaevolvemath}.  Success is therefore measured by the best solution found during the exploration, rather than the average quality of all attempts.

\begin{figure}[t]
\centering
\includegraphics[width=\textwidth]{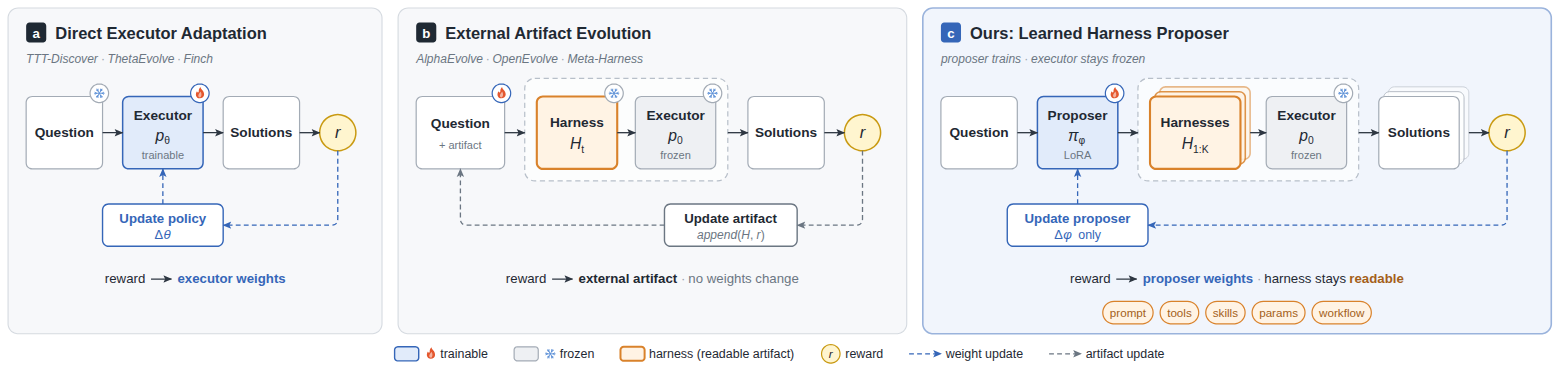}
\caption{
\textbf{Where is test-time reward retained?}
\textbf{(a)}~\emph{Executor weights}: the policy that solves the task is
also the policy being trained~\citep{tttdiscover,thetaevolve}.
\textbf{(b)}~\emph{Proposer context}: the executor stays frozen and a fixed
proposer re-reads an archive or harness accumulated over rounds~\citep{alphaevolve,openevolve,metaharness}.
\textbf{(c)}~\emph{Proposer weights (ours)}: a separate proposer is
trained to write complete harnesses for a frozen executor, so adaptation is
amortized into weights while remaining expressed as inspectable artifacts. \looseness=-1
}
\label{fig:taxonomy}
\end{figure}

Large language models have tackled these tasks through two representative routes as shown in Figure~\ref{fig:taxonomy}.
TTT-Discover~\citep{tttdiscover} performs reinforcement learning at test
time~\citep{ttt,ttrl}, directly updating the executor---the model that solves the
task---from its search outcomes guided by the reward.  This adapts the model to solve the problem directly, but requires repeated sampling and training, with 25,600 rollouts per problem
in its reported configuration~(50 rounds $\times$ 512 rollouts).  Moreover, the learned behavioral adaptation remains encoded in the executor's parameters, making individual changes difficult to isolate and limiting their direct transfer to a different or stronger executor. Artifact 
evolution methods~\citep{funsearch,alphaevolve,openevolve,shinkaevolve,ye2026simpletes} instead keep the executor and harness fixed---adaptation occurs through the feedback artifacts supplied in context. Artifacts are inspectable and reusable, allowing subsequent attempts to build on previous discoveries. However, adaptation within the search occurs through the artifacts supplied in context: the model must infer useful revisions based on its in-context capabilities~\citep{reflexion,selfrefine},
without parameter updates from these outcomes. 
In long-horizon discovery, successive rounds produce increasingly extensive histories. When more of this history is incorporated into the input, accumulating more experience does not reliably translate into more effective exploration~\citep{liu2024lost,hsieh2024ruler}.

To address the above limitations, we introduce \method{}, which updates the proposer's parameters to generate and refine task-specific harnesses while keeping the executor frozen~(as shown in Figure~\ref{fig:one}). 
Our approach takes inspiration from human researchers, who improve not only the solutions they explore, but also the harness through which they explore them. During the exploration, humans may develop reusable problem-solving skills, revise construction methods, build specialized tools, or reorganize the experimental workflow to improve exploration efficiency. On a circle-packing problem, for example, a
researcher might adopt a boundary-first construction strategy, write tools
that check pairwise overlaps and boundary violations or compute the largest
radii a fixed set of centers admits, and organize geometric diagnostics to guide subsequent refinements.  In an agentic system, these changes correspond to adapting the \emph{complete harness} that governs the executor's behavior~\citep{sweagent,weng2026harness,metaharness}. These harness-level adaptations shape an entire subsequent sequence of attempts, rather than only the next candidate solution.  \looseness=-1

\method{} makes such harness adaptation a learned process as shown in Figure~\ref{fig:overview}. At each round, unlike existing methods that adapt only part of the harness---a prompt or context playbook~\citep{opro,dspy,gepa,zhang2026agentic}, a code
patch installed at predefined lifecycle hooks of a fixed
runtime~\citep{harnessr1}, or a fixed four-module protocol (memory, planning,
action, and tool/skill orchestration) that draws on a given tool and skill
registry~\citep{zhang2026jit}---our proposer model generates \emph{complete harnesses} comprising system prompts, skills, tools, middleware, and execution-control parameters.
Each generated harness guides the frozen executor through a budgeted discovery rollout from a shared incumbent solution. Improvements in the best valid solution provide rewards, which update the proposer's parameters via our carefully designed reinforcement learning methods, while only the selected best solution, the incumbent harness, and a compact experience digest are retained for the next round.
Across rounds, discovery experience changes the proposer's policy itself, enabling reliable improvement via learning. Meanwhile, the resulting harness remains explicit and separate from the executor's weights: it can be inspected, revised, and deployed with a stronger executor, providing a mechanism for weak-to-strong transfer~\citep{burns2024weak}.

Implementing \method{} presents
two challenges.
First, \textit{harness-level credit assignment is indirect and noisy}. The proposer generates a harness rather than a final solution, so its reward is observed only through a separate, stochastic executor rollout.
 Even when a fixed solution has a deterministic score, the same harness can
produce different trajectories and yield different best solutions. The observed reward therefore reflects both harness quality and executor sampling variability. 
We develop a discovery-oriented RLOO-based update~\citep{kool2019,rloo}
with gap-normalized rewards, leave-one-out advantages without
variance normalization~\citep{drgrpo}, max-sharpening~\citep{bond,infalign}, and filtering
of uninformative groups~\citep{dapo}, designed to limit noise amplification
while accommodating heterogeneous reward scales and
best-of-group selection.
Second, \textit{harness optimization must guard against reward
hacking and information leakage}. Higher evaluator scores do
not necessarily reflect better solutions: an agent may exploit
flaws in the evaluation process~\citep{specgaming,metr2025} or use information unavailable
under the intended task protocol~\citep{kapoor2023leakage}. We therefore design a new critic component to address this challenge.

We evaluate \method{} on fourteen discovery problems spanning mathematical
constructions~\citep{alphaevolve,alphaevolvemath}, systems
benchmarks~\citep{barbarians}, GPU kernels~\citep{tttdiscover}, heuristic programming
contests~\citep{ahc,alebench}, astrodynamics, and biology. With a frozen
Qwen3.5-9B executor~\citep{qwen35}, \method{} improves on OpenEvolve with the
same executor on all eleven benchmark tasks and holds the best open-weight
sub-10B result on nine, ahead of the discovery-fine-tuned
Finch-9B~\citep{finch} on every task by up to $91.7\%$. Scaling the proposer and the frozen
executor to Qwen3.8-27B~\citep{qwen38} improves on \emph{eleven} of fourteen tasks,
and transferring the learned harnesses to GPT-5.5 matches or exceeds
the strongest reported system of any backbone on \emph{nine} of fourteen tasks and
comes within $1\%$ on four more. Our ablations show where the gains come
from: at matched rollouts, routing reward into the proposer's weights reaches
better results earlier than updating the executor or the proposer's context;
wider search per round yields late breakthroughs; and our RLOO objective with
max-sharpening outperforms standard GRPO and a frozen proposer. Together,
these results show that the search procedure itself can be learned at test
time with a small frozen model and carried over to stronger executors.

%% file: subtex/related.tex
\section{Related Work}
\label{sec:related}

\para{Self-evolution by adapting the model.} One family improves an agent
by changing model weights. Test-time training and test-time RL adapt weights
per instance or per problem~\citep{ttt,ttrl}, while test-time scaling spends more inference compute~\citep{s1}; for discovery,
TTT-Discover~\citep{tttdiscover} and ThetaEvolve~\citep{thetaevolve} fine-tune
the executor on its own search rollouts and hold strong published $\le$10B
records, while discovery fine-tuning amortizes the skill across tasks by
training on evolution-trajectory corpora, as in the Finch models we use as a
parameter-matched comparator~\citep{finch}. Closest to us, HASE~\citep{hase} and SIA~\citep{sia} co-evolve weights
\emph{and} harness, which couples the improver to the improved and requires
update access to the executor; \method{} freezes the executor and trains a
separate proposer policy.

\para{Self-evolution by editing the harness.} A second family keeps the
model frozen and revises an external artifact. Evolutionary program search
around frozen LLMs produces state-of-the-art discovery programs:
FunSearch~\citep{funsearch}, AlphaEvolve~\citep{alphaevolve,alphaevolvemath},
open reimplementations and successors~\citep{openevolve,shinkaevolve},
heuristic evolution~\citep{eoh}, equation discovery~\citep{llmsr}, reward
design~\citep{eureka}, and multi-agent and meta-evolution
systems~\citep{qu2026coral,liu2026evox}, evaluated on algorithm-engineering
suites~\citep{alebench} and the systems problems of~\citet{barbarians};
research agents automate the surrounding pipeline~\citep{aiscientist}. A
parallel line lets a frozen model rewrite its own code or scaffold:
STOP~\citep{stop}, the Darwin G\"odel Machine~\citep{dgm}, self-improving
coding agents~\citep{sica,godelagent}, capability-synthesizing
generalists~\citep{alita}, and a survey of the space~\citep{selfevolvesurvey}.
A third optimizes the text or structure around the model: prompt
search~\citep{ape,opro,promptbreeder,evoprompt,metaprompting}, prompt
compilation~\citep{dspy,mipro}, textual gradients~\citep{textgrad,trace},
reflective evolution~\citep{gepa}, automated agent and workflow
design~\citep{adas,aflow,agentsquare,agentsymbolic}, whole-harness
editing~\citep{metaharness,selfharness}, and in-context
self-correction~\citep{voyager,reflexion,selfrefine}. Across this family the proposer's \emph{weights are fixed}, so design
knowledge is re-derived per task rather than accumulated. We train the
proposer across rounds, and its action space includes executable
tool code: unlike tool creation that abstracts functions from solution
traces~\citep{latm,creator,craft,trove}, the tools here are emitted by a
\emph{learned} policy and admitted through a leak-limiting review chain
(\S\ref{sec:method-safety}).

%% file: subtex/method.tex
\section{\method{}}
\label{sec:method}
\method{} learns the discovery procedure by training a harness
proposer while keeping the executor frozen. The proposer specifies how the
executor searches through proposed harnesses; the executor conducts budgeted rollouts to generate candidate solutions and interacts with evaluators to obtain scores for them.
Feedback from these evaluators updates the proposer and the search state
retained between rounds. We first formulate the discovery objective, then
describe the harness representation, proposer update, iterative search, and
validation protocol.

\para{Problem formulation and framework overview.}
A task $\tau$ provides a description $d_\tau$, a seed program $x_0$ with a
marked editable region, and a black-box evaluator $m_\tau$; the seed program
and evaluator are taken from~\citet{finch}. Each rollout may call the
evaluator at most $B$ times. An agent pairs a frozen language-model executor
$M_0$ with an executable harness $H$ that governs how $M_0$ interacts with
the task. Each outer round starts from a search state: the incumbent program
$x^\star$ (the best found so far, initially $x_0$) with score
$b_\tau=m_\tau(x^\star)$, the incumbent harness $H^\star$, and an experience
digest $D$.

Given this context, the proposer $\pi_\phi$ samples $K$ harness
specifications. Each round keeps a single incumbent harness $H^\star$, but
proposals are dealt round-robin between $H^\star$ and one
diversity-preserving runner-up parent: the best candidate within 2\% of the
top score that adds a new component, or else the second-highest scorer. This
keeps components from non-winning candidates alive for another
round~\citep{gepa,dgm}. In the first round all $K$ proposals share the
initial harness. Each specification is merged into its parent, validated,
and compiled into a candidate harness $H_k$; all candidates start from the
same program $x^\star$. Only the proposer's LoRA adapter $\phi$ is
trained~\citep{lora}; $M_0$ is never updated. A rollout of $M_0$ under $H_k$
edits the program and calls the evaluator on programs
$x_{k,1},\dots,x_{k,T}$ ($T\le B$); its score is the best valid score it
finds, $s_k=\max_{t\le T} m_\tau(x_{k,t})$, where invalid programs are
skipped and a rollout without any valid program counts as failed. The round
keeps the best solution found, so the proposer's objective is the best score
in the group, never below the incumbent's:
$J_K(\phi)=\mathbb{E}_{\pi_\phi}[\max(b_\tau,s_1,\dots,s_K)]$. The update in
\S\ref{sec:method-rl} is a group-relative surrogate that favors high-scoring
proposals, not an unbiased gradient of $J_K$.\looseness=-1

\begin{figure}[t]
\centering
\includegraphics[width=\textwidth]{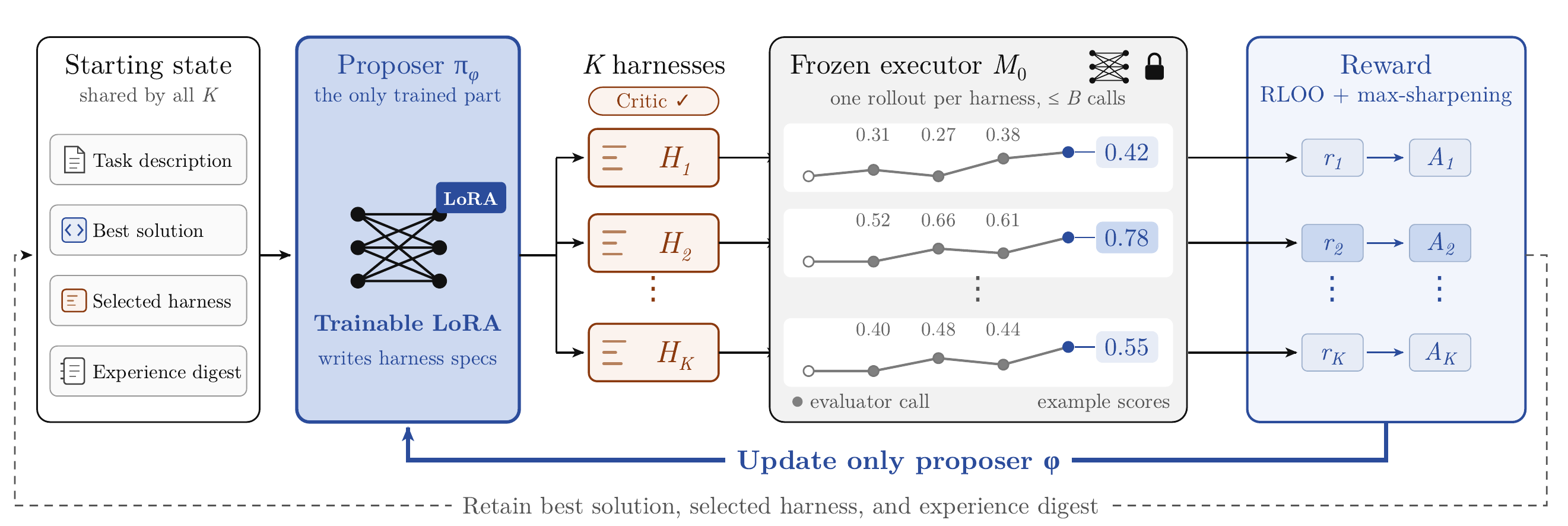}
\caption{\textbf{Overview of \method{}.} A trainable proposer generates $K$
candidate harnesses from a shared task state. Accepted harnesses guide a
frozen executor through budgeted discovery rollouts. Their outcomes provide
group-relative rewards for updating the proposer; selected solutions,
harnesses, and an experience digest condition the next round. Executor
parameters remain fixed throughout.}
\label{fig:overview}

\end{figure}

\para{Executable harness representation.}
What the proposer can learn is bounded by what a harness can express.
Existing systems represent it narrowly: prompt optimizers rewrite
instructions~\citep{promptbreeder,dspy,gepa}, reflective agents append notes
to the context~\citep{reflexion,zhang2026agentic}, and Harness-R1 patches a
fixed runtime without adding executable tools~\citep{harnessr1}. These change
what the executor is told but not what it can do, such as adding an analysis
tool or restarting a stalled search. We instead represent the harness as an
executable program built on NexAU~\citep{nexau} with five editable components
(Figure~\ref{fig:anatomy}; Appendix~\ref{app:h2spec}): a \emph{system prompt}
for strategy and workflow, \emph{skills} for reusable procedures,
\emph{tools} including proposer-written Python functions, \emph{middleware}
hooks that run between executor steps (e.g., budget reminders, restarts after
stalls), and \emph{control parameters} for sampling and iteration limits. The
proposer edits a private copy of $H^\star$ and cannot touch the task program,
evaluator, executor weights, or shared runtime; unchanged fields are
inherited, so a partial specification still yields a complete harness. Each
episode ends with \texttt{validate\_harness}, which checks schemas, mounts,
and generated code (\S\ref{sec:method-safety}), and
\texttt{submit\_harness}; accepted specifications are compiled and hashed to
remove duplicates within the group. The result is a standalone artifact that
can be inspected or run with another frozen executor (\S\ref{app:crossmodel}).

\subsection{Learning the harness proposer}
\label{sec:method-rl}

A harness is judged only through executor rollouts. 
Even with a deterministic evaluator, the same harness can lead a stochastic
executor to different solutions. An observed score therefore mixes the
quality of a proposed procedure with downstream sampling variability.
Our RLOO-based update~\citep{rloo,kool2019} must assign credit under this
indirect feedback while accounting for shrinking absolute gains and an
objective that retains the best discovery rather than the average attempt.

\looseness=-1

\para{Reward.}
As search approaches a strong reference,
meaningful improvements can become small in absolute units. With an
unnormalized leave-one-out advantage, these smaller score differences carry less policy-gradient weight even when they close the same fraction of the remaining gap. 
A valid candidate is rewarded by how much its rollout score $s_k$ improves on
the incumbent score $b_\tau$, relative to the remaining gap to a published
reference score $u_\tau$ (listed in the appendix), or to $|b_\tau|$ once the
incumbent has reached it:

\begin{equation}
g_k=\frac{s_k-b_\tau}{Z_\tau},\qquad
Z_\tau=
\begin{cases}
u_\tau-b_\tau, & b_\tau<u_\tau-\epsilon,\\
|b_\tau|+\epsilon, & \text{otherwise}.
\end{cases}
\label{eq:gap}
\end{equation}

A small remaining gap can make a large gain produce an extreme normalized
reward. We smoothly cap positive credit below the reference, while retaining graded penalties for rollouts that underperform the incumbent: \looseness=-1

\begin{equation}
r_k=
\begin{cases}
3\tanh(g_k/3), & b_\tau<u_\tau-\epsilon,\ g_k>0,\\
\max\{-1,\,g_k\}, & b_\tau<u_\tau-\epsilon,\ g_k\le 0,\\
\operatorname{clip}(g_k,-1,5), & \text{otherwise},
\end{cases}
\label{eq:gap-normalized-reward}
\end{equation}
with $\epsilon=10^{-9}$. Below the reference, gains are smoothly capped, losses are linear and
floored at $-1$, and matching the incumbent earns zero; invalid, failed, or
repaired candidates receive $-1$. The reference $u_\tau$ is used only for the
reward and is never shown to the executor or written into a harness.

Proposals inherit different harness parents, so a cross-parent
comparison could reward an inherited advantage rather than the proposed revision. We compute advantages within groups $\mathcal G$ sharing
a parent and prompt: one group of size $K_g=K$ in the first round, then
two groups of size $K_g=K/2$. Within each group, valid rewards spanning
less than $\delta=10^{-3}$ are replaced by their mean to avoid crediting
near-tied outcomes. Denoting the resulting rewards by $\widetilde r_k$,
we compute the leave-one-out advantage
$
    A_k^{\mathrm{LOO}}=
    \widetilde r_k-\frac{1}{K_g-1}
        \sum_{j\in\mathcal G\setminus\{k\}}\widetilde r_j.
$
We omit division by the group's reward standard deviation, which can turn
small, potentially noisy differences into order-one
advantages~\citep{drgrpo,grpo}. The RLOO term thus retains the task-relative
magnitude established by gap normalization. All-valid tied groups are
excluded from training~\citep{dapo}; groups receiving identical status
penalties are assigned zero advantages and skipped before computing a
valid-reward range or temperature. Mixed groups preserve the contrast
between tied valid rewards and status penalties.  To additionally favor
the strongest outcomes retained by the discovery objective, we augment
RLOO with a centered softmax sharpening term~\citep{bond,infalign}: \looseness=-1

\begin{equation}
    A_k=(1-\alpha)A_k^{\mathrm{LOO}}
       +\alpha\left[
          \operatorname{softmax}
            (\widetilde{\mathbf r}_{\mathcal G}/T_{\mathcal G})_k
          -\frac{1}{K_g}\right],
    \quad
    T_{\mathcal G}=\max\{\Delta_{\mathcal G}/4,\epsilon\},
    \label{eq:advantage}
\end{equation}
where $\alpha=0.3$, $\widetilde{\mathbf r}_{\mathcal G}$ contains the
group's processed rewards, and $\Delta_{\mathcal G}$ is its valid-reward
range. The two terms serve complementary purposes: RLOO preserves the
magnitude of relative progress, while range-scaled sharpening emphasizes
within-group ranking without reducing feedback to a hard winner label.
Near-tie processing suppresses differences below the threshold, but does
not distinguish genuine gains from executor noise. The combined advantage
is a discovery-oriented heuristic and, like standard GRPO~\citep{grpo}, not an unbiased gradient estimator.

\para{Proposer update.}
The executor generates the solutions, but the proposer generates the action being optimized. We therefore apply $A_k$ only to the proposer-generated
tokens in episode $k$, masking tool observations and all executor tokens.
With $\rho_{k,t}(\phi)$ the ratio of current to sampling-policy token
probabilities and $\mathcal T_k$ the retained token set, we minimize the
clipped surrogate~\citep{ppo} with a
KL penalty toward the frozen base proposer $\pi_{\mathrm{ref}}$:

\begin{equation}
\mathcal{L}(\phi)
  = -\frac{1}{N}\sum_{k}\sum_{t\in\mathcal T_k}
      \min\!\big(\rho_{k,t}A_k,\
      \operatorname{clip}(\rho_{k,t},1-\epsilon_c,1+\epsilon_c)A_k\big)
    +\beta\,\mathcal{L}_{\mathrm{KL}},
\label{eq:proposer-surrogate}
\end{equation}
where $N=\sum_k|\mathcal T_k|$,
$\epsilon_c$ is the framework
default, and $\mathcal{L}_{\mathrm{KL}}$ is the low-variance estimator
$\pi_{\mathrm{ref}}/\pi_\phi-\log(\pi_{\mathrm{ref}}/\pi_\phi)-1$ with
$\beta=0.1$. 
Clipping and the KL anchor constrain the change made from a small batch of
indirect feedback. Zero-advantage episodes are omitted; if no training tokens
remain, the optimizer step is skipped without interrupting state retention.
The persistent rank-64 LoRA adapter is trained for three epochs per round
with learning rate $10^{-5}$ and batch size 8. Episodes are truncated
to 18k tokens while preserving the final submission. The updated adapter
then proposes the next round's harnesses, closing the learning loop without
changing $M_0$.

\subsection{Experience retention and harness validation}

\para{Incumbent and bounded context.}
At the end of each round, the best valid program in the group and its harness
become the new incumbent if they strictly improve on its score
(Figure~\ref{fig:overview}; Appendix~\ref{app:training}), and the next round starts
from them rather than from the seed. Solution quality therefore accumulates,
and rewards measure progress beyond the current starting point. Under a fixed
evaluator the incumbent score never decreases, but this follows from retention
and does not by itself show that the proposer improved. Unlike reflective
methods~\citep{reflexion,gepa}, we do not feed the proposer the history of
earlier candidates. Such a history grows every round, raises cost, and buries
informative outcomes in a long context that models use
poorly~\citep{liu2024lost}. Feedback that drifts across rounds would also make
within-group reward differences reflect context rather than proposer quality.
The proposer instead reads a fixed-size \emph{experience digest}, rebuilt each
round: the incumbent harness and program (truncated), the seed and current best
scores, and a stall flag. Experience across rounds is thus carried by the
digest and the proposer's weights. Proposals from the same parent share this
digest and differ only in sampling seed.\looseness=-1

\para{Harness validation.}
\label{sec:method-safety}
Evaluator scores cannot tell real improvements from exploits of the evaluation
interface~\citep{skalse2022,specgaming,denison2024,baker2025,impossiblebench},
so every generated tool is validated before rollouts. The \emph{critic} in
Figure~\ref{fig:overview} is this validation pipeline, not a learned value
function. Tools access the task only through a restricted interface that
withholds the evaluator implementation, ground truth, and network access, and
they must pass static checks and a sandboxed self-test. A failing tool gets at
most two repair attempts, screened for evaluator or target-score content, and
is dropped if it still fails. These controls
describe the evaluated protocol, not an independently verified guarantee.

%% file: subtex/experiment.tex
\section{Experiments}
\label{sec:experiments}

\subsection{Evaluation Design}
\label{sec:exp-setup}

\para{Task suites.} The parameter-matched comparison with open-weight
systems (Table~\ref{tab:open-under-10b}) uses the
eleven tasks for which Finch-9B reports results~\citep{finch}: five
\emph{mathematical construction} tasks (\textbf{Erd\H{o}s}, \textbf{AC1},
\textbf{AC2}, \textbf{CP26},
\textbf{Hadamard})~\citep{alphaevolve,alphaevolvemath,openevolve}; two
\emph{heuristic-contest programming} tasks (\textbf{AHC039}, \textbf{AHC058})
scored under the official AtCoder 150-case protocol~\citep{ahc,alebench}; and
four \emph{systems algorithm design} tasks (\textbf{EPLB}, \textbf{PRISM},
\textbf{LLM-SQL}, \textbf{Txn}) from the ADRS suite~\citep{barbarians}. The ablation (Table~\ref{tab:ablation}) adds one astrodynamics task (\textbf{V2}) and one
biological denoising task (\textbf{scRNA})~\citep{ye2026simpletes,tttdiscover},
for thirteen tasks. The comparison with all prior methods
(Table~\ref{tab:frontier-transfer}) further includes \textbf{TriMul}, the GPU MODE outgoing Triangle
Multiplicative Update kernel benchmark, scored by geometric-mean latency
across seven fixed input shapes on an NVIDIA H100 (lower is
better)~\citep{tttdiscover,ye2026simpletes}, for fourteen tasks. TriMul~(H100), V2 and scRNA have no
external comparator under the open-model scale rule, and LLM-SQL and Txn are
shown for disclosure and reported with caveats.

\input{tables/frontier}

\para{Evaluation protocol.} Following the default setting of Finch~\citep{finch}, every task provides a description, a
seed program with a delimited editable region, a black-box evaluator, and a
hard per-rollout evaluator-call budget. Tables report native metrics and mark
their optimization direction. The configured executor interface exposes
neither evaluator code nor ground truth (\S\ref{sec:method-safety}).

\para{Implementation and reporting unit.} The principal executor is
frozen Qwen3.5-9B~\citep{qwen35} served with vLLM~\citep{vllm}; the scale
comparison uses frozen Qwen3.8-27B~\citep{qwen38}. The standard configuration runs 20 outer rounds and samples $K{=}32$~(Qwen3.5-9B) or $K{=}8$~(Qwen3.8-27B) harnesses per round; for each
harness, the executor runs a single solution trajectory and may call the
evaluator at most $B{=}20$~(Qwen3.5-9B) or $B{=}5$~(Qwen3.8-27B) times within it; the proposer is a persistent LoRA adapter~\citep{lora} trained with
our proposed advantage (Eq.~\ref{eq:advantage}) for three epochs per round at
batch size 8 under a KL anchor. Tables report the selected valid outcome of
each single search campaign, not an across-campaign mean; reruns of a selected
program characterize execution noise only.  \looseness=-1

\subsection{Results}
\label{sec:exp-results}

\input{tables/sota}

\para{Open-weight $\leq$10B regime.}
Table~\ref{tab:open-under-10b} compares \method{}-9B with every open-weight
system of at most 10B parameters that reports these tasks. With the same
frozen Qwen3.5-9B executor, \method{}-9B improves on OpenEvolve on all eleven
tasks, e.g., from 1.173 to 2.636 on CP26 and from 134.5 to 713.6 on AHC058. It
also improves on the discovery-fine-tuned Finch-9B on all eleven, by up to
91.7\% on Hadamard and 36.2\% on CP26. It holds the best $\leq$10B value on
nine tasks, including Hadamard, where its 0.921 is 60\% above the next best
(0.576, ThetaEvolve), and a tie at 2.635983 on CP26 with the three
executor-updating systems. It trails on two: AC2, where those three systems
reach 0.946--0.947 versus 0.934, and Txn, where Finch-4B reaches 4{,}762
versus 4{,}255. OpenEvolve with proprietary executors (Claude Opus 4.6, Gemini 3 Pro, GPT-5; Table~\ref{tab:open-under-10b}) reports results on six of these tasks. \method{}-9B matches or exceeds the best of them on five, and falls behind only on Txn. This suggests that a learned harness can close much of the gap between a small open-weight executor and far larger proprietary ones, without updating the executor.
On AC2, the reported scores of ThetaEvolve and TTT-Discover are higher than \method{}-9B. However, we find that this comparison is confounded by the initial program: on AC2, both methods start from a hand-designed program that scores far above ours, whereas on the other tasks the initial programs also differ but their initial scores are close to ours. Since the initial program's score strongly affects subsequent evolution, we further discuss the effect of initial-program differences on discovery in \S\ref{app:seed-program}. \looseness=-1

\para{Adaptation stages.}
Table~\ref{tab:ablation} isolates the contribution of each stage of proposer learning.
In-context harness editing improves on the fixed initial harness on twelve
of the thirteen tasks (all except PRISM), and training the proposer's weights
improves further over in-context adaptation on eleven, ties on AHC039, and is
slightly lower only on EPLB ($-0.47\%$). This highlights the importance of training the
proposer's weights rather than relying on in-context adaptation alone.

\para{Model scale.}
We scale up the proposer and the executor by replacing Qwen3.5-9B with Qwen3.8-27B. Because the
27B model is more expensive to run, we reduce the number of harnesses sampled
per round from 32 to 8 and the evaluator calls per rollout from 20 to 5. Even
with this smaller budget, the 27B model
improves over the 9B baseline on ten of the thirteen tasks, ties on CP26 and
PRISM, and is lower only on Hadamard (Table~\ref{tab:ablation}). Counting TriMul
(1261.4 to 1096.4~$\mu$s; Table~\ref{tab:frontier-transfer}), it improves on
eleven of the fourteen tasks.

\para{Weak-to-strong transfer.}
Because the learned harness is explicit and separate from the executor's
weights, it can be moved to a stronger executor without retraining. In the
spirit of weak-to-strong generalization~\citep{burns2024weak}, we take the
harnesses learned with Qwen3.5-9B and Qwen3.8-27B, replace the executor with
GPT-5.5, and continue search (\method{}-Transfer).
Table~\ref{tab:frontier-transfer} compares both variants with the strongest
reported result on each task, regardless of backbone or method. With
transfer, \method{} matches or exceeds this frontier on nine of fourteen
tasks (seven excluding the quarantined LLM-SQL and Txn), is within $1\%$ on four
more, and remains behind only on EPLB. It sets new best values on AC2
(0.9636 vs.\ 0.9629), AHC039 (571{,}450 vs.\ 567{,}503), AHC058, TriMul
(1094.8 vs.\ 1183.0~$\mu$s), and V2, and
ties the frontier on Hadamard and PRISM. Transfer helps most where the small
executor falls short: with GPT-5.5, the 9B harness raises AHC058 from 713.55
to 849.68 and Hadamard from 0.921 to 0.936. The transferred harnesses encode
search strategies, such as layered construction and filling edge space first,
that are not tied to the executor used during
optimization, so discovery procedures can be reused across executors without
transferring executor weights. Because executor choice and extra search are
part of the transfer procedure, these gains do not isolate a pure transfer
effect; \S\ref{app:crossmodel} replays a fixed harness on a fixed executor
set.\looseness=-1

\begin{figure}[t]
\centering
\includegraphics[width=\textwidth, trim=0 0 0 0, clip]{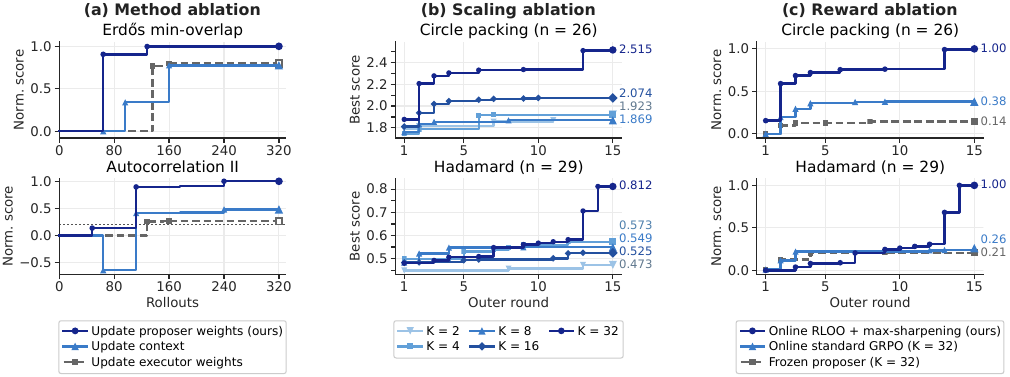}
\caption{\textbf{Ablations}. \textbf{(a)}~Reward routing on
Erd\H{o}s and autocorrelation~II: normalized score versus rollouts spent; every arm
spends 16 rollouts per round (8 proposer and 8 executor rollouts for the
proposer-weight and context arms, 16 executor rollouts for the
executor-weight arm, which has no proposer);~
\textbf{(b)}~Best score over 15 rounds on circle packing and Hadamard for
$K\in\{2,\dots,32\}$ proposals per round; round cost grows with $K$;~\textbf{(c)}~Proposer objective
(\S\ref{sec:method-rl}), normalized as in (a); the RLOO curve is the $K{=}32$
run in (b).}
\label{fig:ablation}
\end{figure}

\subsection{Ablation Study}

\para{Comparison of reward-update locations.}
A rollout reward can update the executor's weights, the proposer's context, or
the proposer's weights. We compare the three routes on two tasks with the same
base checkpoint, evaluator, initial harness, seed program, and 20 evaluator
calls per executor trajectory. The executor arm is a budget-scaled
TTT-Discover-style reference that keeps the initial harness and updates the
executor with a three-epoch rank-64 LoRA on $K{=}16$ final-edit trajectories
per batch. The proposer arm trains a persistent rank-64 LoRA for three epochs
on $K{=}8$ proposals. The context arm makes no gradient update and adds two
frozen-model analyst calls after the cold round. Because each harness candidate
also costs a proposer rollout, Figure~\ref{fig:ablation}a plots every arm
against total rollouts. A harness round spends 8 proposer and 8 executor
rollouts; the executor arm has no proposer, so it runs 16 executor
trajectories per round, and every arm reaches 320 rollouts after 20 rounds. The proposer
arm reaches the highest endpoint on both tasks and gets there sooner. By 128
rollouts it has reached 1.00 on Erd\H{o}s and 0.90 on autocorrelation~II
(normalized), while the context arm is at 0.34 and 0.41, and the executor arm
is still below the plotted range on Erd\H{o}s and at 0.26 on
autocorrelation~II. Updating the proposer thus turns rollout rewards into
better search procedures early in the run.\looseness=-1

\para{Test-time scaling.}
We study how discovery performance changes with search width
and depth by varying the number of harness proposals per round,
$K\in\{2,4,8,16,32\}$, and running each configuration for
15 outer rounds on CP26 and Hadamard
(Figure~\ref{fig:ablation}b).
The widest configuration, $K=32$, achieves the highest final
score on both tasks: 2.515 on CP26 and 0.812 on Hadamard,
compared with 2.074 and 0.573 for the best narrower configurations,
respectively. It also makes substantial late-stage improvements on both tasks, while the narrower configurations largely plateau. \looseness=-1

\looseness=-1

\para{Reward design.}
Figure~\ref{fig:ablation}c varies only the proposer objective, with the same
frozen executor, initial harness, seed program, budget, and $K{=}32$ over
15 rounds; scores are normalized as in Figure~\ref{fig:ablation}a. Updating
the proposer helps under either objective, and ours helps most: with the
RLOO endpoint at 1.00, GRPO reaches 0.38 on CP26 and 0.26 on Hadamard, and
the frozen proposer 0.14 and 0.21. GRPO's curve is nearly flat after
round~4 on both tasks, whereas RLOO keeps improving. This is consistent with
\S\ref{sec:method-rl}: without division by the group standard deviation,
near-tied groups no longer turn evaluator noise into large advantages, and
max-sharpening concentrates credit on the candidate the round keeps. Each
condition is a single campaign.

\looseness=-1

\begin{figure}[t]
\centering
\includegraphics[width=\textwidth]{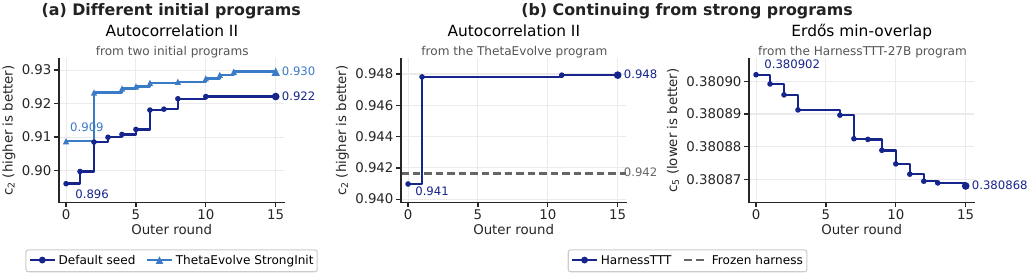}
\caption{\textbf{Search from different initial programs.}
\textbf{(a)}~Autocorrelation~II from the default seed and from the ThetaEvolve
StrongInit program. \textbf{(b)}~Search continued from already-strong
programs: Autocorrelation~II from a ThetaEvolve program and Erd\H{o}s from the
\method{}-27B program. Curves show the incumbent score after each outer
round.}
\label{fig:seed-program}
\end{figure}

\subsection{Effect of the Initial Program}
\label{app:seed-program}

\para{Different initial programs.}
The campaigns in \S\ref{sec:exp-results} all start from the seed program of
each task, so we first ask how the search depends on that starting point.
Figure~\ref{fig:seed-program}a runs \method{}-9B on autocorrelation~II from two
programs: the default seed and the stronger initial program released with
ThetaEvolve (StrongInit)\footnote{\url{https://github.com/ypwang61/ThetaEvolve/blob/main/Results/SecondAutoCorrIneq/programs/StrongInit.py}}. Both runs improve over 15 rounds, from 0.896 to 0.922
and from 0.909 to 0.930. The default-seed run passes StrongInit's starting
score by round~3, and the gap between the two runs narrows from 0.013 to
0.007. It does not close, however, so the final score still depends on the
initial program.

\para{Continuing from strong programs.}
Figure~\ref{fig:seed-program}b asks whether the search still helps when the
starting program is already strong. On autocorrelation~II, starting from a
ThetaEvolve RL@65 program
at 0.941, \method{} reaches 0.948, whereas the frozen initial harness reaches
only 0.942 from the same program. Almost all of this gain arrives in the first
round. On Erd\H{o}s, starting from the best program of the \method{}-27B campaign
(0.380902), the search lowers $c_5$ to 0.380868 and improves in most rounds.
The remaining headroom is small in both cases, and the Erd\H{o}s result is
still above the reported frontier of 0.380859
(Table~\ref{tab:frontier-transfer}). 

\subsection{Cross-Executor Replay of Learned Harnesses}
\label{app:crossmodel}

\begin{figure}[!t]
\centering
\includegraphics[width=0.85\textwidth]{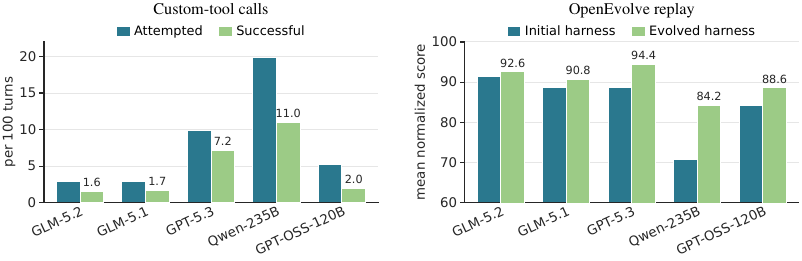}
\caption{\textbf{Cross-executor replay of the 9B harness.} \emph{Left}: custom-tool
calls per 100 turns (eleven tasks, three decode seeds). \emph{Right}: mean
normalized score after 48 OpenEvolve iterations from the initial versus the
evolved harness (one search seed). Point estimates; the aggregate includes the
two quarantined tasks.}
\label{fig:harness-transfer}
\end{figure}

Because the learned harness is explicit, it can be replayed on executors it
was never optimized with. We fix the harnesses learned with Qwen3.5-9B (the
\emph{evolved harness}) and swap the executor for GLM-5.2,
GLM-5.1~\citep{glm5clean}, GPT-5.3~\citep{openai2026gpt53codex},
Qwen-235B~\citep{qwen3}, or
GPT-OSS-120B~\citep{openai2025gptoss120bgptoss20bmodel}, without using any of
their outcomes to train the proposer. For each executor,
OpenEvolve~\citep{openevolve} runs 48 iterations from the initial or the
evolved harness under identical settings, and NexAU~\citep{nexau} records
tool use over three decode seeds. The evolved harness raises the eleven-task
mean normalized score on all five executors by $+1.2$ to $+13.4$ points
($+5.4$ on average; Figure~\ref{fig:harness-transfer}). The gains track
custom-tool use: Qwen-235B calls the generated tools most often (19.9
attempts per 100 turns) and gains the most ($+13.4$), followed by GPT-5.3
(9.9; $+5.7$). The harness thus carries a reusable search procedure that
other executors can execute, without proposer retraining or executor-weight
updates; with one search seed per executor, these gains are descriptive point
estimates.

%% file: tables/frontier.tex
\providecommand{\B}[1]{\bfseries #1}
\providecommand{\U}[1]{\underline{#1}}
\providecommand{\NA}{--}
\providecommand{\dl}[1]{{\itshape\color{ink2} #1}}
\providecommand{\grp}[1]{\multicolumn{15}{c}{\cellcolor{oursbg}\emph{#1}}}

\begin{table}[t]
\centering
\tsz
\setlength{\tabcolsep}{2.8pt}
\renewcommand{\arraystretch}{1.05}
\caption[HarnessTTT-9B versus open-weight systems]{\textbf{HarnessTTT-9B leads
the compared open-weight ($\leq$10B) systems on nine of eleven tasks and
improves on Finch-9B on all eleven.} Bold: best open-weight value per column;
underline: second; \emph{Best}: column wins. Proprietary rows are context
only; the last row is the relative change over Finch-9B in the metric's
direction. $^\dagger$\, Obtained with a different evolution pipeline and shown
for reference only.\protect\footnotemark}
\label{tab:open-under-10b}
\begin{adjustbox}{max width=\textwidth}
\begin{tabular}{lrrrrrrrrrrrc}
\toprule
\textbf{Method}
& \textbf{Erd\H{o}s}$\downarrow$ & \textbf{AC1}$\downarrow$ & \textbf{AC2}$\uparrow$
& \textbf{CP26}$\uparrow$ & \textbf{Hadamard}$\uparrow$
& \textbf{AHC039}$\uparrow$ & \textbf{AHC058 (M)}$\uparrow$
& \textbf{EPLB}$\uparrow$ & \textbf{PRISM}$\uparrow$
& \textbf{LLM-SQL}$\uparrow$ & \textbf{Txn}$\uparrow$ & \textbf{Best} \\
\midrule
Initial program
& 0.495056 & 1.518600 & 0.855800 & 0.959764 & 0.143275
& 534{,}850 & 0.00 & 0.1265 & 21.89 & 0.6856 & 2824.86 & \\
\midrule
\multicolumn{13}{c}{\textit{OpenEvolve, proprietary executors (reference only)}} \\
Claude Opus 4.6
& 0.381880 & \NA & \NA & 2.629300 & \NA
& \NA & \NA & 0.1270 & 26.26 & 0.7160 & 3774.00 & \\
Gemini 3 Pro
& \NA & \NA & \NA & 2.541400 & \NA
& \NA & \NA & 0.1272 & 26.24 & 0.7258 & 4273.50 & \\
GPT-5
& \NA & \NA & \NA & 2.541400 & \NA
& \NA & \NA & 0.1272 & 26.23 & 0.7155 & 4237.30 & \\
\midrule
\multicolumn{13}{c}{\textit{Open-weight executors ($\leq$10B)}} \\
Qwen3.5-2B + OpenEvolve
& 0.381737 & 1.518600 & 0.864600 & 1.253056 & 0.478009
& 546{,}078 & 0.00 & 0.1265 & 21.89 & 0.6856 & 2832.86 & 0/11 \\
Finch-2B  + OpenEvolve
& 0.381346 & 1.518400 & 0.892000 & 1.535134 & 0.400476
& 545{,}256 & 329.36 & \U{0.1269} & 22.26 & 0.6860 & 2949.85 & 0/11 \\
Qwen3.5-4B  + OpenEvolve
& 0.416924 & 1.518600 & 0.880200 & 1.680787 & 0.384332
& 542{,}077 & 0.00 & 0.1266 & 21.89 & 0.6856 & 2732.24 & 0/11 \\
Finch-4B  + OpenEvolve
& 0.386460 & 1.517300 & 0.893300 & 1.806808 & 0.146199
& 551{,}844 & 331.47 & 0.1267 & 22.87 & 0.6857 & \B{4761.90} & 1/11 \\
Qwen3.5-9B + OpenEvolve
& 0.385512 & 1.518600 & 0.880100 & 1.172702 & 0.397184
& 553{,}582 & 134.49 & \U{0.1269} & 22.36 & 0.6858 & 3584.23 & 0/11 \\
Finch-9B + OpenEvolve
& 0.381100 & 1.514100 & 0.912200 & \U{1.936000} & 0.480585
& \U{553{,}759} & \U{525.29} & 0.1265 & \U{23.93} & \U{0.7024} & 3636.36 & 0/11 \\
TTT-Discover$^\dagger$ (Qwen3-8B)
& \U{0.380932} & \U{1.505250} & \B{0.947200} & \B{2.635983} & \NA
& \NA & \NA & \NA & \NA & \NA & \NA & 2/11 \\
ThetaEvolve$^\dagger$  (R1-Qwen3-8B)
& \NA & 1.506803 & \U{0.946900} & \B{2.635983} & \U{0.576400}
& \NA & \NA & \NA & \NA & \NA & \NA & 1/11 \\
Nano-Discover$^\dagger$  (Qwen3-8B)
& 0.380935 & 1.505395 & 0.945766 & \B{2.635983} & \NA
& \NA & \NA & \NA & \NA & \NA & \NA & 1/11 \\
\midrule
\oursrow HarnessTTT-9B (ours)
& \B{0.380919} & \B{1.504680} & 0.933900 & \B{2.635983} & \B{0.921053}
& \B{559{,}534} & \B{713.55} & \B{0.1272} & \B{26.26} & \B{0.7415} & \U{4255.32} & \B{9/11} \\
\dl{$\Delta$ vs.\ Finch-9B (\%)}
& $+$0.05 & $+$0.62 & $+$2.38 & $+$36.2 & $+$91.7
& $+$1.04 & $+$35.8 & $+$0.55 & $+$9.74 & $+$5.57 & $+$17.0 & \\
\bottomrule
\end{tabular}
\end{adjustbox}
\end{table}
\footnotetext{We follow the default evolution pipeline of Sky-Discover:
\url{https://github.com/skydiscover-ai/skydiscover}.}

\begin{table}[t]
\centering
\small
\setlength{\tabcolsep}{2.0pt}
\renewcommand{\arraystretch}{1.05}
\caption{\textbf{Each stage of proposer learning improves the harness on most
tasks, and a larger executor adds a further step on ten of thirteen.} The 9B
rows differ only in how the proposer is updated. Each $\Delta$ row is the
relative change over the row above it, in the metric's direction.
Bold: best per column; underline: second.}
\label{tab:ablation}
\begin{adjustbox}{max width=\textwidth}
\begin{tabular}{lrrrrrrrrrrrrr}
\toprule
\textbf{Condition}
& \textbf{Erd\H{o}s}$\downarrow$ & \textbf{AC1}$\downarrow$ & \textbf{AC2}$\uparrow$
& \textbf{CP26}$\uparrow$ & \textbf{Hadamard}$\uparrow$
& \textbf{AHC039}$\uparrow$ & \textbf{AHC058 (M)}$\uparrow$
& \textbf{EPLB}$\uparrow$ & \textbf{PRISM}$\uparrow$
& \textbf{LLM-SQL}$\uparrow$ & \textbf{Txn}$\uparrow$
& \textbf{V2}$\downarrow$ & \textbf{scRNA}$\uparrow$ \\
\midrule
Initial, 9B
& 0.456591 & 1.518200 & 0.896100 & 1.477767 & 0.360961
& 557{,}225 & 134.49 & 0.1265 & \U{24.02} & 0.6856 & 3610.11
& 3.839529 & 0.641 \\
\addlinespace[2pt]
Context, 9B
& 0.395849 & 1.509800 & 0.926300 & \U{2.370463} & 0.517212
& \U{559{,}534} & 562.82 & \U{0.1278} & 23.04 & 0.7267 & 3906.25
& 3.458023 & 0.701 \\
\dl{\quad $\Delta$ (\%)}
& $+$13.3 & $+$0.55 & $+$3.37 & $+$60.4 & $+$43.3
& $+$0.41 & $+$318.5 & $+$1.03 & $-$4.08 & $+$6.00 & $+$8.20
& $+$9.94 & $+$9.36 \\
\addlinespace[2pt]
\oursrow HarnessTTT-9B
& \U{0.380919} & \U{1.504680} & \U{0.933900} & \B{2.635983} & \B{0.921053}
& \U{559{,}534} & \U{713.55} & 0.1272 & \B{26.26} & \U{0.7415} & \U{4255.32}
& \U{3.446415} & \U{0.709} \\
\dl{\quad $\Delta$ (\%)}
& $+$3.77 & $+$0.34 & $+$0.82 & $+$11.2 & $+$78.1
& 0.00 & $+$26.8 & $-$0.47 & $+$14.0 & $+$2.04 & $+$8.94
& $+$0.34 & $+$1.14 \\
\addlinespace[2pt]
\oursrow HarnessTTT-27B
& \B{0.380902} & \B{1.503778} & \B{0.934868} & \B{2.635983} & \U{0.806604}
& \B{564{,}457} & \B{841.39} & \B{0.1285} & \B{26.26} & \B{0.9676} & \B{4370.00}
& \B{3.430206} & \B{0.714} \\
\dl{\quad $\Delta$ (\%)}
& $+$0.004 & $+$0.06 & $+$0.10 & 0.00 & $-$12.4
& $+$0.88 & $+$17.9 & $+$1.02 & 0.00 & $+$30.5 & $+$2.69
& $+$0.47 & $+$0.71 \\
\bottomrule
\end{tabular}
\end{adjustbox}
\end{table}

%% file: tables/sota.tex
\begin{table}[t]
\centering
\tsz
\setlength{\tabcolsep}{1.5pt}
\renewcommand{\arraystretch}{1.08}
\providecommand{\hsz}{\fontsize{6}{6.8}\selectfont}
\providecommand{\hc}[1]{\multicolumn{1}{c}{\hsz\begin{tabular}[b]{@{}c@{}}#1\end{tabular}}}
\providecommand{\hcc}[1]{\multicolumn{2}{c}{\hsz\begin{tabular}[b]{@{}c@{}}#1\end{tabular}}}
\caption{\textbf{With harness transfer, HarnessTTT matches or exceeds the
reported frontier on nine of fourteen tasks and is within 1\% on four more.}
Baseline columns are the strongest reported system per task family
regardless of backbone. For HarnessTTT, the first column searches
with its own frozen executor and the second moves the learned harness to
GPT-5.5 and continues search. \textbf{Bold}: state of the art per task;
\underline{underline}: second best. TriMul entries are geometric-mean
latencies ($\mu$s) under the GPU MODE kernelbot protocol and software stack,
all re-measured on one H100 in interleaved passes. Entries that agree at the
printed precision are ranked by their unrounded scores. Each transfer column
is a separate search with GPT-5.5 under the learned harness, so executor and
evaluation variance can leave it slightly below the own-executor column on a
few tasks. \looseness=-1}
\label{tab:frontier-transfer}
\resizebox{\textwidth}{!}{%
\begin{tabular}{@{}l*{10}{c}*{4}{>{\columncolor{oursbg}}c}@{}}
\toprule
& \multicolumn{10}{c}{\textbf{Reported systems}} & \multicolumn{4}{c}{\textbf{Our method}} \\
\cmidrule(lr){2-11}\cmidrule(l){12-15}
{\hsz\textbf{Task}}
& \hc{OpenEvolve}
& \hc{AlphaEvolve}
& \hc{Hyra}
& \hc{SimpleTES}
& \hc{TTT-Discover}
& \hc{ShinkaEvolve}
& \hc{CORAL}
& \hc{ALE-Agent}
& \hc{GEPA}
& \hc{LEVI}
& \hcc{HarnessTTT (9B)} & \hcc{HarnessTTT (27B)} \\
{\hsz\textit{backbone}}
& \hc{\textit{multiple}} & \hc{\textit{Gemini}\\\textit{2.0}} & \hc{\textit{Hy3}}
& \hc{\textit{gpt-oss-}\\\textit{120B}} & \hc{\textit{gpt-oss-}\\\textit{120B}}
& \hc{\textit{multiple}} & \hc{\textit{Opus}\\\textit{4.6}} & \hc{\textit{Gemini/}\\\textit{GPT}}
& \hc{\textit{GPT-5/}\\\textit{Gem.\,3}} & \hc{\textit{mixed}}
& \hc{\textit{Qwen3.5-}\\\textit{9B}} & \hc{\textit{transfer} $\to$\\\textit{GPT-5.5}}
& \hc{\textit{Qwen3.8-}\\\textit{27B}} & \hc{\textit{transfer} $\to$\\\textit{GPT-5.5}} \\
\specialrule{\lightrulewidth}{\aboverulesep}{0pt}
\grp{Mathematical discovery} \\
\specialrule{\lightrulewidth}{0pt}{\belowrulesep}
Erd\H{o}s $\downarrow$
& 0.380910 & 0.380924 & \B{0.380859} & \U{0.380868} & 0.380875 & 0.38156 & 0.38089 & \NA & \NA & \NA
& 0.380919 & 0.380923 & 0.380902 & \U{0.380868} \\
AC1 $\downarrow$
& 1.506604 & 1.503170 & \B{1.502850} & 1.503871 & \U{1.502863} & \NA & \NA & \NA & \NA & \NA
& 1.504680 & 1.506805 & 1.503778 & 1.503871 \\
AC2 $\uparrow$
& 0.944900 & 0.961020 & \U{0.962901} & 0.962694 & 0.959180 & \NA & \NA & \NA & \NA & \NA
& 0.933900 & 0.934210 & 0.934868 & \B{0.963589} \\
CP26 $\uparrow$
& \U{2.635983} & \U{2.635983} & \U{2.635983} & \U{2.635983} & \NA & \U{2.635983} & \B{2.6360} & \NA & \NA & \NA
& \U{2.635983} & \U{2.635983} & \U{2.635983} & \U{2.635983} \\
Hadamard $\uparrow$
& \NA & \NA & \NA & \B{0.935673} & \NA & \NA & \NA & \NA & \NA & \NA
& \U{0.921053} & \B{0.935673} & 0.806604 & \B{0.935673} \\
\specialrule{\lightrulewidth}{\aboverulesep}{0pt}
\grp{Algorithm engineering} \\
\specialrule{\lightrulewidth}{0pt}{\belowrulesep}
AHC039 $\uparrow$
& \NA & \NA & \NA & 567{,}503 & 567{,}062 & 558{,}026 & \NA & 550{,}647 & \NA & \NA
& 559{,}534 & \B{571{,}450} & 564{,}457 & \U{569{,}409} \\
AHC058 (M) $\uparrow$
& \NA & \NA & \NA & 849.33 & 848.41 & \NA & \NA & 848.37 & \NA & \NA
& 713.55 & \U{849.68} & 841.39 & \B{850.80} \\
\specialrule{\lightrulewidth}{\aboverulesep}{0pt}
\grp{System performance} \\
\specialrule{\lightrulewidth}{0pt}{\belowrulesep}
EPLB $\uparrow$
& 0.1272 & \NA & \NA & \NA & \NA & 0.1290 & \U{0.1490} & \NA & 0.1445 & \B{0.1523}
& 0.1272 & 0.1294 & 0.1285 & 0.1288 \\
PRISM $\uparrow$
& \B{26.26} & \NA & \NA & \NA & \NA & \B{26.26} & \B{26.26} & \NA & \U{26.23} & \B{26.26}
& \B{26.26} & \B{26.26} & \B{26.26} & \B{26.26} \\
LLM-SQL $\uparrow$
& 0.7258 & \NA & \NA & \NA & \NA & 0.7240 & 0.7310 & \NA & 0.7134 & 0.7985
& 0.7415 & 0.7375 & \B{0.9676} & \U{0.9672} \\
Txn $\uparrow$
& 4{,}273.50 & \NA & \NA & \NA & \NA & 4{,}329.00 & \U{4{,}566.00} & \NA & 3{,}984.10 & 4{,}464.29
& 4{,}255.32 & \B{4{,}651.16} & 4{,}370.00 & 4{,}273.50 \\
\specialrule{\lightrulewidth}{\aboverulesep}{0pt}
\grp{GPU kernel engineering} \\
\specialrule{\lightrulewidth}{0pt}{\belowrulesep}
TriMul ($\mu$s) $\downarrow$
& \NA & \NA & \NA & 1183.0 & 1290.6 & \NA & \NA & \NA & \NA & \NA
& 1261.4 & 1147.3& \U{1096.4} & \B{1094.8} \\
\specialrule{\lightrulewidth}{\aboverulesep}{0pt}
\grp{Astrodynamics and biology} \\
\specialrule{\lightrulewidth}{0pt}{\belowrulesep}
V2 $\downarrow$
& \NA & \NA & \NA & 3.430214 & \NA & \NA & \NA & \NA & \NA & \NA
& 3.446415 & 3.432584 & \U{3.430206} & \B{3.430204} \\
scRNA $\uparrow$
& 0.710 & \NA & \NA & \B{0.740} & 0.730 & \NA & \NA & \NA & \NA & \NA
& 0.709 & \U{0.740} & 0.714 & 0.736 \\
\bottomrule
\end{tabular}%
}

\end{table}

%% file: subtex/conclusion.tex
\section{Conclusion}

We introduced \method{}, which casts test-time discovery as learning the
harness rather than the model: a proposer is trained online to write
executable harnesses for a frozen executor. With a frozen 9B executor,
\method{} improves on OpenEvolve and Finch-9B on all eleven benchmark tasks
and holds the best open-weight ($\leq$10B) value on nine; a 27B model
improves further on eleven of fourteen tasks; and with transfer to GPT-5.5
it matches or exceeds the reported frontier on nine of fourteen
tasks and comes within $1\%$ on four more. Ablations show that, at matched
rollouts, updating the proposer's weights reaches better results earlier than
updating the executor or the context, and that RLOO with max-sharpening
outperforms GRPO and a frozen proposer. Together, these results show that
harness search can itself be learned at test time, complementary to scaling
the model that executes it.
\looseness=-1

%% file: subtex/appendix.tex
\section{Overview}
\label{app:overview}

This appendix provides additional details on \method{}. It is organized as
follows.
\begin{enumerate}[leftmargin=*,itemsep=2pt,topsep=2pt]
    \item \textbf{Appendix~\ref{app:overview-figure}: \method{} on One Page.}
          A one-page summary of the two test-time evolution loops, the
          properties a learned harness provides, the tasks on which it reaches
          the reported frontier, the tools the proposer wrote, and one
          circle-packing trajectory.
    \item \textbf{Appendix~\ref{app:h2spec}: The Harness Spec Language.}
          The fields a proposer may edit (system prompt, skills, tools,
          middleware, and control parameters), their types and limits, and how
          a partial specification is merged over its parent and hashed to
          remove duplicates.
    \item \textbf{Appendix~\ref{app:sdk}: Capability Surface of Generated Tools.}
          The \texttt{ctx} interface available to generated tools, the
          operations it withholds, and the static checks and sandboxed
          self-tests a tool must pass before it is admitted.
    \item \textbf{Appendix~\ref{app:training}: Proposer Training Details.}
          The round configuration (outer rounds, harnesses per round, and
          evaluator budget per rollout), parent selection and advantage
          grouping, the RLOO and max-sharpening hyperparameters, and the LoRA,
          optimizer, and KL settings of the proposer update.
    \item \textbf{Appendix~\ref{app:reference-scores}: Published Reference
          Scores.} The reference score $u_\tau$ used to normalize the reward on
          each task, its source, and its optimization direction; these scores
          enter only the reward and are never shown to the executor.
    \item \textbf{Appendix~\ref{app:discovered-harness}: Discovered Harnesses.}
          Examples of harnesses found by the trained proposer, including the
          generated tools, skills, and middleware, and the search strategies
          they encode.
    \item \textbf{Appendix~\ref{app:routing-case-study}: Audited Reward-Routing
          Case Study.} Reward-routing trajectories on autocorrelation~II under
          the three reward routes, and an artifact-level audit of the events
          behind them.
\end{enumerate}

\section{\method{} on one page}
\label{app:overview-figure}

Figure~\ref{fig:one} summarizes \method{} on a single page. Panel~(a)
contrasts two test-time evolution loops that differ only in what the reward
updates. In executor RL, as in TTT-Discover~\citep{tttdiscover}, the executor
$\theta$ samples $N$ rollouts, each covering one strategy; the reward updates
$\theta$, and the final weights $\theta^\star$ cannot be transferred. In
\method{}, the proposer $\phi$ writes $K$ harnesses---named, readable files for
the prompt, skills, tools, and middleware---that the frozen executor $M_0$
rolls out. A single harness-guided rollout can move through several strategies
(A$\to$B$\to$C). Candidates are scored and gated, the reward updates only
$\phi$, and the best harness $H^\star$ is a transferable package. Panel~(b)
lists the four properties that follow. Exploration happens within a rollout.
Every edit is traceable as a named file. Generated code is kept safe by
schema, scope, and sandbox checks. The harness transfers to other frozen
executors.

Panel~(c) marks the tasks on which the learned harness, with or without
transfer, matches or exceeds the strongest reported system of any backbone
(Table~\ref{tab:frontier-transfer}). It exceeds the frontier on seven tasks: AC2
against Hyra~\citep{hyra2026}; AHC039, AHC058, TriMul, and V2 against
SimpleTES~\citep{ye2026simpletes}; LLM-SQL against
LEVI~\citep{tanveer2026levi}; and Txn against CORAL~\citep{qu2026coral}. It
ties the frontier on Hadamard (SimpleTES) and on PRISM
(OpenEvolve~\citep{openevolve}, ShinkaEvolve~\citep{shinkaevolve}, CORAL, and
LEVI). LLM-SQL and Txn are quarantined. On four more tasks, it is within $1\%$
of the frontier: Erd\H{o}s and AC1 (Hyra), circle packing (CORAL;
EurekAgent~\citep{xin2026eurekagent}), and scRNA (SimpleTES). It remains behind
only on EPLB (LEVI).

Panel~(d) shows the task-specific harnesses the proposer writes on four tasks,
with the executor frozen. Each card gives the goal, the visible effect on the
solution, and the generated tool and skill or middleware: a staggered-packing
constructor and packing guide for circle packing, a basin-and-bottleneck
analyzer with a multi-start strategy for Hadamard, a column-ranking tool and
row-sorting playbook for LLM-SQL, and a conflict-graph builder with graph-first
middleware for transaction scheduling. Panel~(e) follows one circle-packing
($n{=}26$) campaign with a 9B proposer and executor. Named harness edits raise
the sum of radii from 0.960 at the seed to 1.467 in round~1 (broaden the
search), 2.235 in round~5 (test four layouts), 2.278 in round~7 (escape a
plateau), and 2.541 in round~10 (import a warm start). Replayed unchanged on a
frozen GLM-5.2 executor, the same harness reaches 2.635984 in all nine runs,
within $0.001\%$ of the reported frontier
(2.6360 by CORAL~\citep{qu2026coral}; 2.635999 by EurekAgent~\citep{xin2026eurekagent}).

\begin{figure}[h]
\centering
\includegraphics[width=\textwidth]{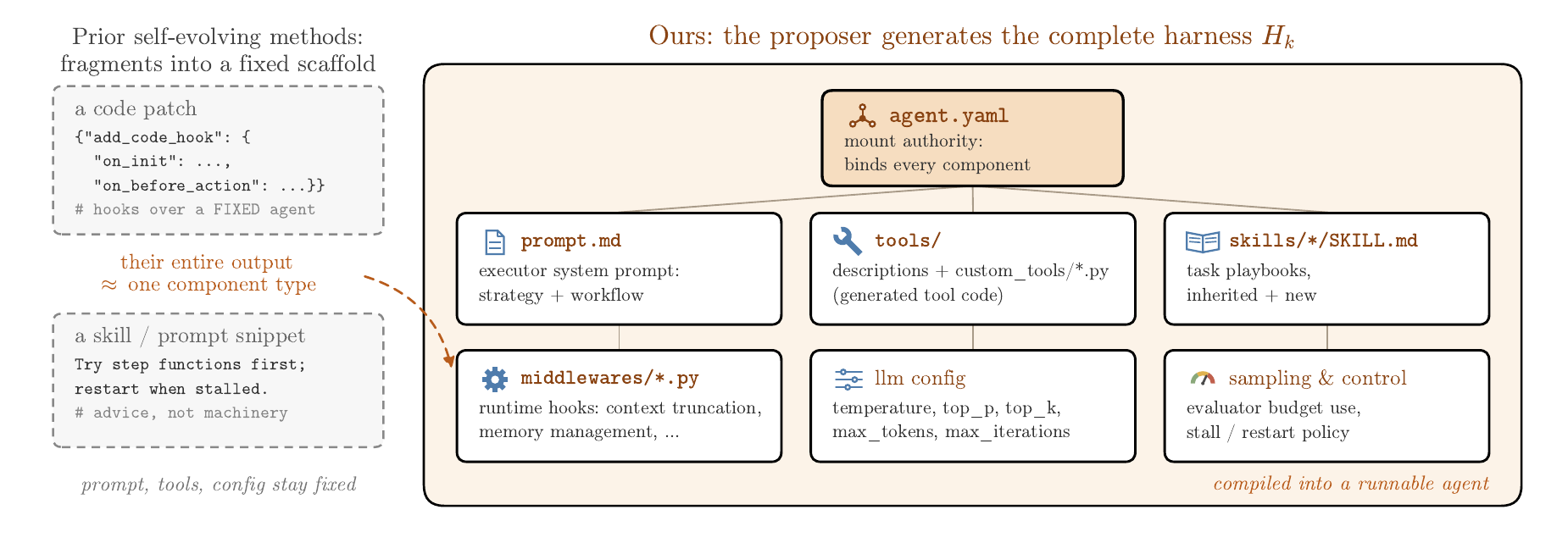}
\caption{\textbf{What a generated harness contains.}  Prior methods emit
fragments---a patch or a skill snippet~\citep{harnessr1,voyager}---into a
fixed scaffold. Our proposer can change every component of the executable
harness, which is bound by \texttt{agent.yaml} and compiled into a runnable
agent.}
\label{fig:anatomy}
\end{figure}

\section{The harness spec language}
\label{app:h2spec}

The proposer's output is a \emph{harness spec}: a declarative document that
fixes every component the frozen executor sees (Figure~\ref{fig:anatomy}).
A candidate spec is a \emph{partial} YAML document merged over its parent
spec. Fields the candidate sets override the parent's, and fields it omits are
inherited. A proposal that rewrites only the skill playbook therefore still
defines a complete harness, and what a candidate changes can be read directly
from the fields it sets. Table~\ref{tab:h2spec} lists the fields.

\begin{table}[h]
\centering
\small
\caption{\textbf{Fields of a harness spec.} The upper block overrides
components the parent already has; the lower block changes the set of
components.}
\label{tab:h2spec}
\begin{tabular}{@{}l l p{0.5\textwidth}@{}}
\toprule
Field & Type & Semantics \\
\midrule
\multicolumn{3}{@{}l}{\emph{Overrides of existing components}} \\
\texttt{system\_prompt}      & text & executor system prompt \\
\texttt{skill\_description}  & text & skill card shown in the executor's tool list \\
\texttt{skill\_body}         & text & skill playbook body \\
\texttt{tool\_descriptions}  & map  & descriptions of the fixed tools \\
\texttt{sampling}            & map  & temperature, top-$p$, top-$k$, max tokens \\
\texttt{agent}               & map  & iteration cap \\
\texttt{middleware}          & map  & budget-reminder onset; stall-restart threshold
                                      and cap; output truncation \\
\midrule
\multicolumn{3}{@{}l}{\emph{Changes to the component set}} \\
\texttt{new\_tools}          & list & $\le 3$; name, description, JSON schema,
                                      Python implementation \\
\texttt{new\_skills}         & list & $\le 2$; additional skill packages \\
\texttt{new\_middlewares}    & list & $\le 2$; name, hook, Python implementation \\
\texttt{remove\_tools}       & list & tools to remove; only optional tools qualify \\
\bottomrule
\end{tabular}
\end{table}

\para{Overriding existing components.}
The upper block rewrites what the executor reads and how it is run. Three text
fields carry the natural-language part of the harness. The system prompt sets
the executor's workflow. The skill package is split into a short card that
advertises the skill in the executor's tool list and a playbook body that holds
the procedure itself. \texttt{tool\_descriptions} rewrites how the fixed tools
are described without touching their code, so the proposer can steer when and
how the executor calls them. \texttt{sampling} and \texttt{agent} set the
decoding parameters and the iteration cap. \texttt{middleware} tunes the
built-in controls: when the budget reminder starts, when a stalled run is
restarted and how many restarts are allowed, and the output-truncation limit.

\para{Changing the component set.}
The lower block adds or removes components. A candidate may add up to three
tools, each with a name, a description, a JSON schema for its arguments, and a
Python implementation. It may also add up to two skill packages, and up to two
middlewares, each attached to a named hook with its own Python implementation.
Tools can be removed only if they are marked optional. These caps bound how much
new code a single candidate can introduce.

\para{Constraints on generated code.}
Generated tool names must match \verb|^[a-z][a-z0-9_]{2,31}$|: they start with
a lowercase letter, contain 3--32 lowercase letters, digits, or underscores, and
may not shadow a reserved name. Each generated implementation is limited to
6{,}000 characters. A generated tool exposes the single \texttt{run(ctx, args)}
entry point and reaches the task only through \texttt{ctx}
(Appendix~\ref{app:sdk}). Before it is mounted, it must pass the static gates
and sandboxed self-test of \S\ref{sec:method-safety}. A tool that still fails
after repair is dropped rather than shipped.

\para{Validation, compilation, and hashing.}
Validation is fail-closed: anything that does not pass is excluded rather than
let through. The proposer calls \texttt{validate\_harness}, which checks
component mounts, field schemas, and generated code, and finishes with
\texttt{submit\_harness} (\S\ref{sec:method}). A deterministic
materializer then merges the candidate over its parent, compiles the result
into a runnable package bound by \texttt{agent.yaml}, and writes a canonical
spec of the effective harness. Within one materializer revision, identical
effective specs yield identical canonical-spec bytes and therefore the same
hash. Provenance metadata stored with the package may still differ, so the hash
identifies the effective harness rather than any particular package. Because
the hash is taken after merging, two candidates in a group that reach the same
effective harness through different partial documents share a hash. The hash
is used to de-duplicate them.

\section{Capability surface of generated tools}
\label{app:sdk}

\begingroup
\setlength{\emergencystretch}{3em}
A generated tool touches the task only through the \texttt{ctx} object passed
to its \texttt{run(ctx, args)} entry point (Table~\ref{tab:ctx}). Through this
interface a tool can read the current and best programs, stage edits, score
them, and read the task's inputs. It cannot see how the scores are computed.

\begin{table}[h]
\centering
\small
\caption{\textbf{The \texttt{ctx} interface available to generated tools.}}
\label{tab:ctx}
\begin{tabular}{@{}l l p{0.42\textwidth}@{}}
\toprule
Group & Method & Behavior \\
\midrule
Program state & \texttt{get\_program()}              & current program \\
              & \texttt{get\_best\_program()}        & best program so far \\
              & \texttt{best\_score()}               & score of the best program \\
Editing       & \texttt{stage\_edit(code)}           & stage an edit to the program \\
Scoring       & \texttt{evaluate()}                  & full evaluation; uses one of the $B$ evaluator calls \\
              & \texttt{probe()}                     & subsampled evaluation, $\sim$8$\times$ cheaper;
                                                       outside $B$ but separately capped \\
              & \texttt{budget\_left()}              & remaining evaluator calls \\
Task inputs   & \texttt{list\_task\_inputs()}        & names of the task inputs \\
              & \texttt{read\_input\_sample(name, n)} & $n$ records of a named input \\
              & \texttt{read\_input\_df(name, n)}    & the same, as a \texttt{pandas} DataFrame \\
Utilities     & \texttt{scratch\_write/read}         & scratch storage \\
              & \texttt{log}                         & log messages \\
\bottomrule
\end{tabular}
\end{table}

\para{Scoring and budget.}
The two scoring calls differ in cost and accounting. \texttt{evaluate()} runs
the full evaluation and uses one of the $B$ evaluator calls. \texttt{probe()}
scores the program on a subsample at roughly one-eighth the cost. It does not
count toward $B$, but it has its own cap. A tool can therefore screen edits
cheaply and save budgeted calls for promising ones, but the cap stops it from
using probes as unlimited free evaluations.

\para{Isolation.}
Path resolution refuses evaluator and ground-truth locations. The configured
sandbox gives generated tools no network, general-filesystem, or subprocess
access.

\para{Static gate and self-test.}
The static gate admits imports only from \texttt{math}, \texttt{re},
\texttt{json}, \texttt{itertools}, \texttt{functools}, \texttt{collections},
\texttt{heapq}, \texttt{bisect}, \texttt{random}, \texttt{statistics},
\texttt{string}, \texttt{typing}, \texttt{dataclasses}, \texttt{numpy}, and
\texttt{pandas}. It also applies the forbidden-access checks of
\S\ref{sec:method-safety}. A tool that passes the gate then runs a self-test
against a mock context under \texttt{RLIMIT\_CPU} 20\,s and \texttt{RLIMIT\_AS}
6\,GB.

\para{Verifiability.}
The public artifact does not currently include the trusted runtime, sandbox,
gate, scrubber, localization guard, or served-model request logs. These
settings therefore describe the evaluated protocol and cannot yet be verified
independently.
\par
\endgroup

\section{Proposer training details}
\label{app:training}
\label{app:infra}

Table~\ref{tab:hparams} gives the configuration of both proposer-training
campaigns: the Qwen3.5-9B campaigns of \S\ref{sec:exp-results}, and the
Qwen3.8-27B campaign whose harnesses Appendix~\ref{app:discovered-harness}
lists. The two campaigns share the search, reward, and update recipe. They
differ only in model scale, proposals per round, the evaluator budget per rollout,
and the reference score $u_\tau$ (Appendix~\ref{app:reference-scores}). The paragraphs below explain how
the pieces fit together.

\begin{table}[h]
\centering
\small
\newcommand{\hpboth}[1]{\multicolumn{2}{>{\centering\arraybackslash}p{\dimexpr0.64\textwidth+2\tabcolsep\relax}}{#1}}
\newcommand{\hpna}{\textcolor{gray}{---}}
\renewcommand{\arraystretch}{1.08}
\caption{\textbf{Proposer-training hyperparameters.} Values centered across
both columns are shared; \hpna: not recorded for the 9B campaigns.}
\label{tab:hparams}
\begin{tabular}{@{}>{\raggedright\arraybackslash}p{0.27\textwidth}
                >{\centering\arraybackslash}p{0.32\textwidth}
                >{\centering\arraybackslash}p{0.32\textwidth}@{}}
\toprule
\textbf{Parameter} & \textbf{27B campaign} & \textbf{9B campaigns} \\
\midrule
\multicolumn{3}{@{}l}{\textbf{Search}} \\
Outer rounds per task      & \hpboth{20} \\
Proposals per round $K$    & 8 & 32 \\
Trajectories per round     & $2K = 16$ & $2K = 64$ \\
Parent beam                & \hpboth{incumbent + best candidate within 2\% of the top score that adds a new component, else the second-highest scorer} \\
Advantage group size $K_g$ & \hpboth{$K$ in the first round, $K/2$ afterwards} \\
Round acceptance           & \hpboth{$\ge 6$ proposals validate (\texttt{min6})} \\
Program promotion          & \hpboth{strict score improvement} \\
\midrule
\multicolumn{3}{@{}l}{\textbf{Executor rollout}} \\
Executor (frozen)          & Qwen3.8-27B & Qwen3.5-9B \\
Serving                    & vLLM, TP\,2\,$\times$\,4 replicas, one 8-GPU node & \hpna \\
Context window             & 49{,}152 tokens & \hpna \\
Evaluator calls $B$        & 5 & 20 \\
Time per evaluation        & 420\,s & 600\,s \\
Wall-clock limit           & \hpboth{3\,h} \\
\midrule
\multicolumn{3}{@{}l}{\textbf{Proposer episode}} \\
Proposer                   & Qwen3.8-27B + LoRA & Qwen3.5-9B + LoRA \\
Harness                    & \hpboth{fixed two-tool proposer harness, workspace preloaded} \\
Agent iterations           & \hpboth{$\le 12$} \\
Sampling                   & \hpboth{temperature 1.0, top-$p$ 0.95} \\
Max generated tokens       & \hpboth{60{,}000} \\
Tool repair                & \hpboth{$\le 2$ iterations by the same frozen model} \\
\midrule
\multicolumn{3}{@{}l}{\textbf{Reward and advantage}} \\
Normalized gain $g_k$      & \hpboth{$(s_k-b_\tau)/(u_\tau-b_\tau)$} \\
Shaped reward $r_k$        & \hpboth{$3\tanh(g_k/3)$ if $g_k>0$, else $\max(-1,g_k)$} \\
Tie threshold $\delta$     & \hpboth{$10^{-3}$} \\
Sharpening weight $\alpha$ & \hpboth{0.3} \\
Softmax temperature $T$    & \hpboth{$\max(\mathrm{range}(r)/4,\,10^{-9})$} \\
All-tied groups            & \hpboth{excluded ($A_k=0$)} \\
\midrule
\multicolumn{3}{@{}l}{\textbf{Proposer update}} \\
Framework                  & \hpboth{slime, FSDP backend, 8 GPUs} \\
LoRA                       & \hpboth{rank 64, $\alpha_{\mathrm{LoRA}}=128$; \texttt{q,k,v,o,gate,up,down}} \\
Optimizer                  & \hpboth{Adam, $\beta_1=0.9$, $\beta_2=0.98$, weight decay 0} \\
Learning rate              & \hpboth{$10^{-5}$, constant} \\
Batch                      & \hpboth{micro 1; global 8} \\
Epochs per round           & \hpboth{3} \\
KL penalty                 & \hpboth{0.1 (\texttt{low\_var\_kl}) to the frozen base} \\
Episode length             & \hpboth{$\le 18{,}000$ tokens, terminal-preserving} \\
\bottomrule
\end{tabular}
\end{table}

\paragraph{Round structure.}
Each task runs for 20 outer rounds. In each round the proposer samples $K$
harnesses, and each candidate harness is rolled out once by the frozen
executor and compared against the shared incumbent score. A round therefore
launches $2K$ agent trajectories (one proposer episode and one candidate
rollout per candidate): 16 in the 27B campaign and 64 in the 9B campaigns.
To limit the cost of the larger executor, the 27B campaign samples $K=8$
proposals rather than 32. It also gives each rollout $B=5$ evaluator calls
of up to 420\,s, where the 9B campaigns give $B=20$ calls of up to 600\,s. Every rollout is capped at 3\,h of wall-clock time. A
round is accepted only if at least six of its proposals validate (the
\texttt{min6} gate). The same frozen model may repair a generated tool that
fails validation at most twice.

\paragraph{Parent selection and grouping.}
Programs and harnesses are promoted separately. The program registry is a
strict single-program ratchet: a rollout's program replaces the incumbent only
if it strictly improves the score. The incumbent \emph{harness} is promoted
only when the round's best candidate exceeds the historical best; displaced
parents are archived. From the second round on, the $K$ proposal slots are dealt round-robin to
a beam of two parents: the promoted or retained incumbent, and the best
candidate within 2\% of the top score that adds a new component (or else
the second-highest scorer). An advantage group
contains the proposals that share one sampling condition (parent spec, prompt,
and budget hash). Hence $K_g = K$ in the first round and $K/2$ afterwards, which is 8
and then 4 for the 27B campaign, and 32 and then 16 for the 9B campaigns.

\paragraph{Rewards and advantages.}
Both campaigns compare each candidate with the shared incumbent score
$b_\tau$. Below the reference,
\begin{equation*}
g_k = \frac{s_k - b_\tau}{u_\tau - b_\tau},\qquad
r_k = \begin{cases} 3\tanh(g_k/3) & g_k > 0,\\ \max(-1,\, g_k) & g_k \le 0,\end{cases}
\end{equation*}
as in Eqs.~\ref{eq:gap} and~\ref{eq:gap-normalized-reward}. Here $u_\tau$ is the reference
score of Appendix~\ref{app:reference-scores}. The positive branch compresses
large gains smoothly, and the negative branch keeps regressions linear down to
$-1$. Invalid, failed, or repaired candidates receive $r_k=-1$, so they cannot
earn credit for a gain. Among the creditable
candidates, rewards within $\delta=10^{-3}$ of each other are collapsed to
their mean. The advantage then follows Eq.~\ref{eq:advantage} with group size
$K_g$:
$A_k = 0.7\,A^{\mathrm{LOO}}_k + 0.3\,\big(\sftm(r/T)_k - 1/K_g\big)$ with
$T=\max(\mathrm{range}(r)/4,\,10^{-9})$. A group whose candidates are all
valid and tied carries no signal and is excluded ($A_k=0$), as is a group in
which every candidate receives the same status penalty; mixed groups keep the
contrast between tied valid rewards and status penalties.

\paragraph{Proposer update.}
Training runs on slime with the FSDP backend on 8 GPUs, for three epochs per
round over all trainable episodes of the round. Each episode is converted into
a replay of its full multi-turn trajectory, and the loss mask covers assistant
turns that are closed by a later user or tool message. Episodes are truncated
to 18{,}000 tokens in a terminal-preserving way: tool observations are clipped
first, and then the longest fitting suffix that starts at an assistant action
is kept. The rewarded \texttt{submit\_harness} action therefore always survives
truncation. The loss is a clipped policy-gradient objective with the
precomputed advantage $A_k$ broadcast over the masked tokens. slime's own
reward normalization is disabled, consistent with the absence of variance
scaling in Eq.~\ref{eq:advantage}. A KL penalty of 0.1
(\texttt{low\_var\_kl}) anchors the policy to the frozen base. The adapter is
cumulative: each round loads only the weights of the previous round's adapter
(\texttt{-{}-load-weights-only}). Every round's checkpoint is retained. The
merged checkpoint is served as the next round's proposer, while the executor
stays at the base checkpoint.

\section{Published reference scores}
\label{app:reference-scores}

The reward of \S\ref{sec:method-rl} divides a candidate's gain over its
baseline (the incumbent score $b_\tau$; Appendix~\ref{app:training}) by the baseline's remaining gap to a
task reference $u_\tau$. The reference therefore sets the scale of credit on
each task; it is not a target that the proposer or the executor ever sees.

In the 27B campaign, $u_\tau$ is the largest of four fixed values: the seed
score, the best human value, the published $\le$10B SOTA, and the Finch-9B
score. All four are expressed in the evaluator's combined score.
Table~\ref{tab:reference-scores} lists the resulting references. The 9B campaigns use
the published $\le$10B SOTA alone (Table~\ref{tab:open-under-10b}), which
changes the reference on the four tasks where the best human value is
largest: Erd\H{o}s, Hadamard, AHC039, and AHC058. When $u_\tau$ does not
exceed the baseline, the remaining gap is not positive. The reward then falls
back to the relative gain over $|b_\tau|$, clipped to $[-1, 5]$. The reference
enters the reward only. It is never written into a harness, a prompt, or a
tool, and a leak guard rejects any proposal or repair whose text contains it.

\begin{table}[h]
\centering
\small
\setlength{\tabcolsep}{6.5pt}
\renewcommand{\arraystretch}{1.1}
\newcommand{\refgrp}[1]{\multicolumn{7}{@{}l}{\textit{#1}}\\\addlinespace[1pt]}
\caption{\textbf{Reference scores $u_\tau$ of the 27B campaign.} $x$ is the
evaluator's combined score (higher is better); $f$ maps it to the native metric
of Table~\ref{tab:frontier-transfer}.}
\label{tab:reference-scores}
\begin{tabular}{@{}l l r r r l l@{}}
\toprule
Task & Metric & $f(x)$ & $u_\tau$ & $f(u_\tau)$ & Source & Backbone \\
\midrule
\refgrp{Mathematical discovery}
Erd\H{o}s $\downarrow$ & $c_5$        & 0.380922\,/\,$x$ & 0.999987  & 0.380927  & Best human   & ---         \\
AC1 $\downarrow$       & $c_1$        & 1.505293\,/\,$x$ & 1.000029  & 1.505250  & TTT-Discover & Qwen3-8B    \\
AC2 $\uparrow$         & $c_2$        & 0.896280\,$x$    & 1.056813  & 0.947200  & TTT-Discover & Qwen3-8B    \\
CP26 $\uparrow$        & sum of radii & 2.635\,$x$       & 1.000373  & 2.635983  & ThetaEvolve  & R1-Qwen3-8B \\
Hadamard $\uparrow$    & det.\ ratio  & $x$              & 0.935673  & 0.935673  & Best human   & ---         \\
\addlinespace[3pt]\refgrp{Algorithm engineering}
AHC039 $\uparrow$      & total score  & 225{,}000\,$x$   & 2.519987  & 566{,}997 & Best human   & ---         \\
AHC058 (M) $\uparrow$  & total score  & 450\,$x$         & 1.883722  & 847.67    & Best human   & ---         \\
\addlinespace[3pt]\refgrp{System performance}
EPLB $\uparrow$        & balancedness & $x$              & 0.1269    & 0.1269    & OpenEvolve   & Finch-2B    \\
PRISM $\uparrow$       & combined     & $x$              & 23.93     & 23.93     & OpenEvolve   & Finch-9B    \\
LLM-SQL $\uparrow$     & combined     & $x$              & 0.7024    & 0.7024    & OpenEvolve   & Finch-9B    \\
Txn $\uparrow$         & combined     & $x$              & 4{,}761.9 & 4{,}761.9 & OpenEvolve   & Finch-4B    \\
\bottomrule
\end{tabular}
\end{table}

\section{Discovered harnesses}
\label{app:discovered-harness}

This section lists the final incumbent harness that the trained Qwen3.8-27B
proposer produced on two of its evolution tasks, circle packing and Erd\H{o}s
minimum overlap (20 rounds each). The final harness is the package that the
last round's collector retained as the parent.
Table~\ref{tab:final-harnesses} summarizes the two harnesses. For each
task we then give the lineage of the final harness and excerpts of the compiled
artifacts on disk: the effective spec with long bodies replaced by pointers,
the generated tool code, the generated skill playbooks, and, for circle
packing, the policy sections the proposer wrote into the executor's system prompt.

Both harnesses keep the base configuration: sampling at $T{=}0.7$, top-$p$
0.95, top-$k$ 20, and at most 8{,}192 tokens; 36 executor iterations; and the
base middleware thresholds (budget reminder from 3 evaluations left, stall
restart after 8 stalls, at most 2 restarts, tool output clipped at 8{,}000
characters). The proposer's edits went into the prompt, the skills, and the
tools.

\begin{table}[H]
\centering
\footnotesize
\setlength{\tabcolsep}{4.5pt}
\renewcommand{\arraystretch}{1.15}
\caption{\textbf{The two final 27B harnesses.} Best program: the campaign's
retained program in native units (Table~\ref{tab:reference-scores}) and the
round that produced it.}
\label{tab:final-harnesses}
\begin{tabular}{@{}l c l c c r >{\raggedright\arraybackslash}p{0.266\textwidth}@{}}
\toprule
Task & Harness & Spec hash & Tools & Skills & Best program & Proposer's prompt policy \\
\midrule
CP26 $\uparrow$        & r13/c06 & \hname{128394fe5159501a} & 2 & 2 & 2.635983 (r18) & preflight--repair--rank loop; transform as comparator \\
Erd\H{o}s $\downarrow$ & r19/c01 & \hname{f940301924f28c6a} & 2 & 5 & 0.380902 (r19) & five-evaluation funding rule; hard-max-first reporting \\
\bottomrule
\end{tabular}
\end{table}

\subsection{Circle packing ($n{=}26$): a verifier and a transform in front of every evaluation}

The lineage (Table~\ref{tab:lineage-circle}) shows the harness being assembled
in the order a human engineer might choose. A feasibility checker comes first
(round 0), then a construction playbook that tells the executor how to use it
(round 5), then a transform that proposes repaired geometry (round 6), and
finally a first-moves policy that stops the executor from discarding the
incumbent's optimizer (rounds 8 and 13). The checker,
\ctool{packing\_feasibility} (Listing~\ref{lst:packing-feasibility}), uses a
water-filling relaxation to return the largest radii consistent with the walls
and every pair, so the executor can rank layouts and repair the listed breaches
without spending an evaluation. Seven further generated tools proposed in
rounds 1--19 (\ctool{packing\_repair}, \ctool{packing\_variant\_generator},
\ctool{local\_packing\_optimizer}, \ctool{packing\_family\_scan}, \ldots) were
rolled out but are not part of the final harness.

\begin{table}[H]
\centering
\footnotesize
\renewcommand{\arraystretch}{1.15}
\caption{\textbf{Circle packing: lineage of the final harness.} Harness score
(combined; native $=2.635\,x$) after each round, and the generated components
of the round's best candidate: \textcolor{sky}{tools}, \textcolor{heather}{skills},
and \textit{prompt edits}. Shaded: the round that produced the final harness.}
\label{tab:lineage-circle}
\begin{tabular}{c >{\raggedright\arraybackslash}p{0.604\textwidth} c r}
\toprule
Round & Components of the round's best candidate & Promoted & Harness score \\
\midrule
0      & \ctool{packing\_feasibility} & \lgy & 0.964486 \\
1--4   & \ctool{packing\_repair}; \ctool{packing\_variant\_generator} + skill; \ctool{local\_packing\_optimizer} & \lgn & 0.964486 \\
5      & \cskill{circle-packing-construction} & \lgy & 0.990712 \\
6      & \ctool{packing\_transform} & \lgy & 0.997374 \\
7      & \cprompt & \lgy & 0.998137 \\
8      & \cskill{dense-constructor-first-moves} & \lgy & 0.999731 \\
9--12  & \ctool{packing\_family\_scan}; \cprompt[prompt edits] & r12 only & 0.999731 \\
\rowcolor{oursbg}
13     & \cskill{dense-constructor-first-moves} (updated) & \lgy & 1.000371 \\
14--19 & \ctool{packing\_family\_probe}, \ctool{packing\_refine}, \ctool{packing\_family\_diversifier}, \cskill{topology-family-enumeration} & \lgn & 1.000371 \\
\bottomrule
\end{tabular}
\end{table}

\begin{lstlisting}[style=harnessyaml,caption={\textbf{Circle packing: effective spec of the final harness.} Long bodies are replaced by pointers.},label={lst:circle-spec}]
schema: h2spec/1.0
skill_description: Iteratively optimize a program's EVOLVE-BLOCK to maximize an automatic evaluator score,
  under a fixed evaluation budget. Use for construction, algorithm-speed, and heuristic discovery tasks
  scored by combined_score (higher is better) through the edit_solution / evaluate_solution / finish tools.
tool_descriptions:
  edit_solution: 'Change the code inside the # EVOLVE-BLOCK region. Two accepted formats: (1) PREFERRED
    - one or more targeted SEARCH/REPLACE diff blocks, each matching the current program exactly: ''<<<<<<<
    SEARCH'' / exact original lines / ''======='' / new lines / ''>>>>>>> REPLACE''. Do not rewrite the
    whole region for a small change. (2) A full rewrite - the complete new EVOLVE-BLOCK body as plain
    code (fenced ```python is allowed; the markers are optional). The fixed entry function and imports
    outside the block are preserved automatically. After editing, call evaluate_solution to score it.'
  evaluate_solution: Run the current program through the task evaluator and report its score. Returns
    combined_score (higher is better), validity, any error, your best score so far, and how many evaluations
    remain. Use the feedback to improve the code, then evaluate again. The best-scoring version is kept
    automatically. Your number of evaluations is limited.
  probe_solution: Cheaply score the CURRENT program on SUBSAMPLED data (first ~2000 rows of any dataset).
    Fast and does NOT consume the evaluation budget, but the score is approximate and NOT comparable to
    full evaluate_solution scores - use probes to rank your own variants when full evaluation is slow,
    then confirm the best one with evaluate_solution.
  finish: End the session. Call this when your evaluation budget is exhausted or you cannot improve the
    best score further. The best-scoring program is submitted automatically.
sampling:
  temperature: 0.7
  top_p: 0.95
  top_k: 20
  max_tokens: 8192
agent:
  max_iterations: 36
middleware:
  budget_reminder_from_left: 3
  stall_after: 8
  max_restarts: 2
  long_tool_output_max_chars: 8000
system_prompt: <9550 chars; see prompt.md>
skill_body: <2388 chars; base discovery-optimization skill>
new_tools:
- name: packing_feasibility
  description: 'Deterministic geometry checker for a candidate circle packing: given 2D centers (and optional
    radii), returns the maximal feasible radii under wall+pair constraints, a validity flag, the specific
    wall/pair violations with breach magnitudes, the sum of radii (the scored quantity), and a coarse
    radius histogram to detect candidate-family collapse. Call it on the exact centers you intend to submit
    before every evaluate_solution; it does not consume evaluation budget.'
  input_schema:
    type: object
    properties:
      centers:
        type: array
        description: List of [x, y] pairs for every circle in the candidate packing. Each coordinate must
          be a finite number. Use the exact centers you intend to submit.
        items:
          type: array
          items:
            type: number
          minItems: 2
          maxItems: 2
        minItems: 2
        maxItems: 200
      radii:
        type: array
        description: Optional. Explicit per-circle radii, same length as centers. When omitted, the tool
          computes the maximal feasible radii for the given centers.
        items:
          type: number
          minimum: 0
        minItems: 2
        maxItems: 200
    required:
    - centers
    additionalProperties: false
  implementation_py: <6587 chars; see packing_feasibility.py>
- name: packing_transform
  description: Deterministic packing generator/repairer. Given current centers (and optional radii), it
    pushes interior-slack circles toward their nearest wall, separates overlapping pairs along their join
    line, then re-solves the max-radius feasibility LP (wall + pair + r>=0). Returns transformed centers,
    maximal feasible radii, resulting sum_radii, validity, residual violations, and a coarse family_key
    for diversity. It is a TRANSFORM (new geometry), distinct from packing_feasibility which only diagnoses.
    Costs no evaluation budget.
  input_schema:
    type: object
    properties:
      centers:
        type: array
        description: List of [x, y] pairs for every circle in the candidate packing (typically 26). Each
          coordinate must be a finite number in [0,1].
        items:
          type: array
          items:
            type: number
          minItems: 2
          maxItems: 2
        minItems: 2
        maxItems: 200
      radii:
        type: array
        description: Optional. Explicit per-circle radii, same length as centers. When omitted, the tool
          solves for maximal feasible radii on the transformed centers.
        items:
          type: number
          minimum: 0
        minItems: 2
        maxItems: 200
    required:
    - centers
    additionalProperties: false
  implementation_py: <5209 chars; see packing_transform.py>
new_skills:
- name: circle-packing-construction
  description: 'Geometry playbook for maximizing the sum of 26 circle radii in a unit square: verify every
    candidate with the packing_feasibility tool before any evaluation, repair from its violation list,
    push slack circles toward walls, and switch structural families on histogram/score plateau.'
  body: <4408 chars; see circle-packing-construction.SKILL.md>
- name: dense-constructor-first-moves
  description: 'Circle-packing first-moves playbook for n=26 in a unit square: compare a keep-and-refine
    of the incumbent constructor against at most one alternate dense template, preflight both with the
    feasibility tool, and spend the limited evaluation budget on the higher-verified candidate.'
  body: <2840 chars; see dense-constructor-first-moves.SKILL.md>
\end{lstlisting}

\begin{lstlisting}[style=harness,caption={\textbf{Circle packing: task policy the proposer wrote into the executor's system prompt} (excerpt of \hname{prompt.md}).},label={lst:circle-policy}]
# Circle-packing task policy (n=26 in a unit square)

Scores the **sum of radii** of 26 circles in a unit square. Hard constraints:
every circle inside the square; no two overlap. A violation => validity 0.

**Preflight before every `evaluate_solution`:** call `packing_feasibility` on
your exact centers; only evaluate when `valid=true` and `sum_radii` exceeds
your best. Costs no evaluation budget.

**No post-hoc clipping** (`np.clip(centers, 0.01, 0.99)`): it silently moves
circles and can create overlaps. Place centers already inside the square.

**Do NOT discard a working optimizer-based constructor** (SLSQP/local
refinement over a seeded lattice + max-radius LP) just to emit a hand-written
literal table; a refined seed usually out-scores a from-scratch layout.
Treat `packing_transform` as a **comparator**, not a mandate: adopt its output
only when its `sum_radii` strictly beats your best -- its wall-push can break
tight triangular contacts and lower the sum. Compare >= 2 construction families
(keep-and-refine vs. one different row/shell/corner template), promote by the
tool's `sum_radii`, confirm the winner, and reserve the last evaluation for
consolidating the highest-`sum_radii` valid candidate.
\end{lstlisting}

\begin{lstlisting}[style=harnesspy,caption={\textbf{Circle packing: core of the generated tool \hname{packing\_feasibility}} (round 0).},label={lst:packing-feasibility}]
def _maximal_radii(centers):
    """Largest radii satisfying r_i <= wall_i and r_i + r_j <= d_ij.

    Water-filling relaxation: start at wall distances, then for every pair
    whose constraint is violated, reduce the larger of the two radii to
    restore feasibility. Repeat until stable or the cap is hit.
    """
    n = len(centers)
    radii = [min(c[0], 1.0 - c[0], c[1], 1.0 - c[1]) for c in centers]
    radii = [max(r, 0.0) for r in radii]
    pairs = []
    for i in range(n):
        for j in range(i + 1, n):
            dx = centers[i][0] - centers[j][0]
            dy = centers[i][1] - centers[j][1]
            d = math.sqrt(dx * dx + dy * dy)
            if d > 0.0:
                pairs.append((i, j, d))
    for _ in range(_MAX_ITERS):            # 200 sweeps, ~13k pair checks for n=26
        changed = False
        for i, j, d in pairs:
            s = radii[i] + radii[j]
            if s > d + _EPS:
                if radii[i] >= radii[j]:
                    new_i = max(0.0, d - radii[j])
                    if radii[i] > new_i + _EPS:
                        radii[i] = new_i; changed = True
                else:
                    new_j = max(0.0, d - radii[i])
                    if radii[j] > new_j + _EPS:
                        radii[j] = new_j; changed = True
        if not changed:
            break
    return radii

def run(ctx, args):
    centers = _parse_centers(args.get("centers"))
    ...
    radii = _parse_radii(args.get("radii"), n)
    if radii is None:
        radii = _maximal_radii(centers); radii_source = "computed_maximal"
    wall, pairs_bad = _violations(centers, radii)
    valid = (len(wall) == 0) and (len(pairs_bad) == 0)
    return {"ok": True, "n_circles": n, "radii_source": radii_source,
            "sum_radii": round(sum(radii), 9), "valid": valid,
            "wall_violations": wall[:20], "pair_violations": pairs_bad,
            "violation_count": len(wall) + len(pairs_bad),
            "radius_histogram": _histogram(radii),
            "hint": ("If valid=false, fix the listed violations: for a pair, move the "
                     "circle with the larger radius away along the line joining the two "
                     "centers; for a wall breach, pull the circle inward. Then re-run "
                     "this tool before calling evaluate_solution."
                     if not valid else
                     "Packing is feasible; sum_radii is the quantity the evaluator rewards.")}
\end{lstlisting}

\begin{lstlisting}[style=harness,caption={\textbf{Circle packing: generated skills.} The mandatory core loop of \hname{circle-packing-construction} (round 5) and the first-moves rule of \hname{dense-constructor-first-moves} (rounds 8 and 13).},label={lst:circle-skills}]
## Core loop (mandatory order)                      -- circle-packing-construction
1. Propose.  After every edit_solution, extract the exact list of 26 [x, y]
   centers the code will emit.
2. Preflight.  Call packing_feasibility with those exact centers (omit radii so
   it computes maximal feasible radii). This costs no evaluation budget.
3. Read.  valid, sum_radii, wall_violations, pair_violations (sorted by breach
   magnitude), radius_histogram.
4. Repair if invalid.  Fix the largest 1-2 breaches per pass: pair breach ->
   move the larger circle away along the join line; wall breach -> pull inward.
   Stop after 3 repair passes -- a still-invalid layout is structurally weak.
5. Rank without evaluating.  Several center sets, ranked by the tool's
   sum_radii; promote at most the top 1-2 to evaluate_solution.
6. Evaluate only valid, pre-ranked layouts.
Family switch (triggered): two consecutive evaluate_solution results gain
< 0.5%, or the radius_histogram is nearly single-bin while sum_radii is
plateaued -> change structure, not coordinates (rows/shells, asymmetric radii,
corner treatment).

## Core rule                                          -- dense-constructor-first-moves
Keep-and-refine is the default first move. Do NOT rewrite a working
optimizer-based constructor (SLSQP/local refinement over a seeded lattice
followed by a max-radius LP solve) into a hand-written literal center table.
1. Keep the incumbent core; make the first edit a targeted improvement to it.
2. One alternate arm for diversity: at most ONE structurally different template.
3. Preflight both with packing_feasibility; rank by sum_radii.
4. Promote the winner; never evaluate an invalid layout.
Budget: most of the session preflighting and comparing, at most 2-3
evaluations on distinct finalists, the last one reserved for consolidation.
\end{lstlisting}

\subsection{Erd\H{o}s minimum overlap: a family ranker, a spectral diagnostic, and a funding rule}

For its first four rounds, the Erd\H{o}s harness gained one skill per round:
objective and projection, a preservation-first method gate, candidate
diversity, and evaluation allocation. Two tools then made these skills
executable. \ctool{overlap\_spectral\_diagnostic} (round 4) recomputes the
task's linear hard-max overlap $c_5 = \max_k \int h(x)(1-h(x+k))\,dx$ for one
candidate and names the binding shifts. \ctool{block\_family\_sweeper}
(round 5; updated in rounds 14, 16, and 19) enumerates a bounded family of
$0/1$ block constructions and ranks them by the same objective. Both tools
compute the \emph{linear} overlap with a zero-padded FFT
(Listing~\ref{lst:block-sweeper}), and the proposer's tool description warns
the executor against the circular-FFT ``illusion'' of the seed program. The
final round's update of the sweeper coincided with the program that beats the
best human construction ($c_5 = 0.380902 < 0.380927$).

\begin{table}[H]
\centering
\footnotesize
\renewcommand{\arraystretch}{1.15}
\caption{\textbf{Erd\H{o}s: lineage of the final harness}, as in
Table~\ref{tab:lineage-circle}. Native $c_5 = 0.380922/x$. $^{\dagger}$The
parent's score dipped in round~18.}
\label{tab:lineage-erdos}
\begin{tabular}{c >{\raggedright\arraybackslash}p{0.604\textwidth} c r}
\toprule
Round & Components of the round's best candidate & Promoted & Harness score \\
\midrule
0      & \cskill{erdos-min-overlap} & \lgy & 0.988310 \\
1      & \cskill{erdos-min-overlap-method} & \lgy & 0.999422 \\
2      & \cskill{erdos-candidate-diversity} & \lgy & 0.999528 \\
3      & \cskill{c5-eval-allocation} & \lgy & 0.999559 \\
4      & \ctool{overlap\_spectral\_diagnostic} & \lgy & 0.999880 \\
5      & \ctool{block\_family\_sweeper} & \lgy & 0.999881 \\
6--7   & \cprompt; \ctool{relaxed\_overlap\_sweeper} (removed later) & \lgy & 0.999894 \\
8      & \cskill{fast-structural-c5-search} & \lgy & 0.999904 \\
9--13  & \ctool{worst\_shift\_repair}; skill removals and re-adds & mixed & 0.999904 \\
14     & \ctool{block\_family\_sweeper} (updated) + \cskill{erdos-min-overlap-method} (re-added) & \lgy & 0.999904 \\
15--16 & \cprompt; \ctool{block\_family\_sweeper} (updated) & \lgy & 0.999925 \\
17--18 & \cskill{c5-budget-focus} replacing \cskill{c5-eval-allocation} & \lgn & 0.833051$^{\dagger}$ \\
\rowcolor{oursbg}
19     & \ctool{block\_family\_sweeper} (updated) & \lgy & 1.000053 \\
\bottomrule
\end{tabular}
\end{table}

\begin{lstlisting}[style=harnessyaml,caption={\textbf{Erd\H{o}s: effective spec of the final harness.}},label={lst:erdos-spec}]
schema: h2spec/1.0
skill_description: Iteratively optimize a program's EVOLVE-BLOCK to maximize an automatic evaluator score,
  under a fixed evaluation budget. Use for construction, algorithm-speed, and heuristic discovery tasks
  scored by combined_score (higher is better) through the edit_solution / evaluate_solution / finish tools.
tool_descriptions:
  edit_solution: 'Change the code inside the # EVOLVE-BLOCK region. Two accepted formats: (1) PREFERRED
    - one or more targeted SEARCH/REPLACE diff blocks, each matching the current program exactly: ''<<<<<<<
    SEARCH'' / exact original lines / ''======='' / new lines / ''>>>>>>> REPLACE''. Do not rewrite the
    whole region for a small change. (2) A full rewrite - the complete new EVOLVE-BLOCK body as plain
    code (fenced ```python is allowed; the markers are optional). The fixed entry function and imports
    outside the block are preserved automatically. After editing, call evaluate_solution to score it.'
  evaluate_solution: Run the current program through the task evaluator and report its score. Returns
    combined_score (higher is better), validity, any error, your best score so far, and how many evaluations
    remain. Use the feedback to improve the code, then evaluate again. The best-scoring version is kept
    automatically. Your number of evaluations is limited.
  probe_solution: Cheaply score the CURRENT program on SUBSAMPLED data (first ~2000 rows of any dataset).
    Fast and does NOT consume the evaluation budget, but the score is approximate and NOT comparable to
    full evaluate_solution scores - use probes to rank your own variants when full evaluation is slow,
    then confirm the best one with evaluate_solution.
  finish: End the session. Call this when your evaluation budget is exhausted or you cannot improve the
    best score further. The best-scoring program is submitted automatically.
sampling:
  temperature: 0.7
  top_p: 0.95
  top_k: 20
  max_tokens: 8192
agent:
  max_iterations: 36
middleware:
  budget_reminder_from_left: 3
  stall_after: 8
  max_restarts: 2
  long_tool_output_max_chars: 8000
system_prompt: <9410 chars; see prompt.md>
skill_body: <2388 chars; base discovery-optimization skill>
new_tools:
- name: overlap_spectral_diagnostic
  description: Cheaply compute the per-shift overlap profile c(k)=int h(x)(1-h(x+k))dx for a single candidate
    step function h on [0,2], given as either explicit 'blocks' or a value array 'h'. Returns the hard-max
    C5 value (the max-k objective), the dominant (worst) shifts, top-2 shift gap, a soft-max temperature
    recommendation, the integral-of-h feasibility check, and head/tail of the profile. Pure deterministic
    diagnostics to rank candidate geometries/parameterizations before committing a full evaluate_solution.
  input_schema:
    type: object
    properties:
      blocks:
        type: array
        description: Optional list of intervals where h=1; each element is {start,end} (in [0,2]) or a
          2-element [start,end]. Provide this OR 'h'.
        items:
          type: object
      h:
        type: array
        description: Optional array of candidate h values in [0,1] over [0,2]; resampled to num_intervals
          if length differs. Provide this OR 'blocks'.
        items:
          type: number
      num_intervals:
        type: integer
        description: Discretization N for the [0,2] grid (default 600).
        default: 600
      top_k:
        type: integer
        description: Number of worst shifts to report (default 3).
        default: 3
    required: []
    additionalProperties: false
  implementation_py: <3923 chars; see overlap_spectral_diagnostic.py>
- name: block_family_sweeper
  description: Deterministically enumerate a bounded family of distinct 0/1 step-function candidates h
    on [0,2] (bump count, spacing mode, width, symmetry), scale each for exact integral 1, and rank them
    by the TASK's LINEAR (non-circular) hard-max C5 = max_k int h(x)(1-h(x+k))dx over k in [0,2] via zero-padded
    FFT (identical method to overlap_spectral_diagnostic, so the ranking matches the true objective).
    Returns ranked candidates with worst shift, top-2 gap, and profile head/tail. Use it to pick which
    block construction to commit to edit_solution/evaluate_solution.
  input_schema:
    type: object
    properties:
      n_bumps:
        type: integer
        description: Number of bumps in the construction family (1-16).
      spacing:
        type: string
        description: 'Spacing mode: ''periodic'' (evenly spaced), ''clustered'' (around center x=1), or
          ''pair'' (mirror-symmetric pairs).'
      bump_width:
        type: number
        description: Base bump width before integral-1 scaling, in [0.01, 1.0].
      symmetry:
        type: string
        description: Reserved symmetry tag for the family ('none', 'mirror'); does not alter generation
          when spacing is 'pair'.
      widths:
        type: array
        description: Optional list of up to 6 alternative widths to sweep within the family.
        items:
          type: number
      n_intervals:
        type: integer
        description: Discretization N for the [0,2] grid (3-2400, default 600).
      max_candidates:
        type: integer
        description: Maximum candidates to generate and rank (default 12).
    required: []
    additionalProperties: false
  implementation_py: <8471 chars; see block_family_sweeper.py>
new_skills:
- name: erdos-min-overlap
  description: 'Playbook for reducing the C5 upper bound in the Erdos minimum-overlap problem: minimize
    max_k int h(x)(1-h(x+k)) dx over step functions h:[0,2]->[0,1] with inth=1. Use whenever editing the EVOLVE-BLOCK
    for this task.'
  body: <3635 chars; see erdos-min-overlap.SKILL.md>
- name: erdos-min-overlap-method
  description: 'Method gate for the Erdos minimum-overlap C5 task: accurate incumbent characterization,
    preservation-first edit gate, and a ranked list of high-leverage polishes to test one per evaluation.'
  body: <4809 chars; see erdos-min-overlap-method.SKILL.md>
- name: erdos-candidate-diversity
  description: Playbook for diversifying step-function candidates in the min-overlap C5 objective so each
    full evaluation covers several distinct basins instead of re-optimizing one hyperparameter variant.
  body: <3590 chars; see erdos-candidate-diversity.SKILL.md>
- name: c5-eval-allocation
  description: 'Allocation rule for the 5-evaluation Erdos C5 budget: confirm the incumbent, allow one
    micro-polish, then fund each remaining evaluation on a distinct candidate family (value scheme, support
    structure, or optimizer), with a 1e-4 switch rule and hard-max-first reporting.'
  body: <2892 chars; see c5-eval-allocation.SKILL.md>
- name: fast-structural-c5-search
  description: 'Workflow for choosing which block construction to invest expensive evaluations in: rank
    families with block_family_sweeper, diagnose worst shifts with overlap_spectral_diagnostic, make one
    structural edit, confirm once.'
  body: <2265 chars; see fast-structural-c5-search.SKILL.md>
\end{lstlisting}

\begin{lstlisting}[style=harnesspy,caption={\textbf{Erd\H{o}s: objective evaluation shared by both generated tools} (from \hname{block\_family\_sweeper}).},label={lst:block-sweeper}]
def _c5_linear_from_blocks(blocks, n_intervals):
    """Hard-max C5 with the TASK's LINEAR (non-circular) correlation.

    c(k) = sum_x h(x) * (1 - h(x+k)) * dx for k in [0, 2], h zero outside [0,2].
    Computed via zero-padded FFT exactly like the proven overlap_spectral_diagnostic,
    so its ranking matches the true objective (not the circular wrap-around value).
    """
    n = int(n_intervals); dx = 2.0 / n
    h = [0.0] * n
    for b in blocks:                      # paint the 0/1 blocks onto the grid
        s = max(0.0, min(2.0, float(b[0]))); e = max(0.0, min(2.0, float(b[1])))
        if e <= s: continue
        i0 = int(math.floor(s / dx)); i1 = min(max(int(math.ceil(e / dx)), i0 + 1), n)
        for i in range(i0, i1): h[i] = 1.0
    integral = sum(h) * dx
    a = np.zeros(n); a[:] = h
    om = 1.0 - a
    af = np.fft.rfft(a, n=2 * n)          # zero padding -> linear correlation
    of = np.fft.rfft(om, n=2 * n)
    corr = np.fft.irfft(af * np.conj(of), n=2 * n)
    c = corr[: n + 1] * dx
    c5 = float(c.max()); worst = int(np.argmax(c))
    csorted = np.sort(c)[::-1]
    gap = float(csorted[0] - csorted[1])  # how isolated the binding shift is
    return c5, round(worst * dx, 6), round(gap, 6), ..., integral

# run(): enumerate {periodic, clustered, pair} x n_bumps x widths, rescale each
# candidate to integral 1, rank by c5_hard_max (lower is better) and return
# {"best": ..., "ranked": [...], "note": "... Use overlap_spectral_diagnostic
#  on your final candidate to confirm before committing to evaluate_solution."}
\end{lstlisting}

\begin{lstlisting}[style=harness,caption={\textbf{Erd\H{o}s: generated skills.} The funding rule of \hname{c5-eval-allocation} (round 3) and the workflow of \hname{fast-structural-c5-search} (round 8).},label={lst:erdos-skills}]
## Funding sequence                                            -- c5-eval-allocation
1. Eval 1 -- confirm baseline.  Fund the incumbent as-is; record its hard-max c5.
2. Eval 2 -- at most one micro-polish (soft-max temperature anneal, short
   near-hard-max refinement, exact integral projection, or higher resolution),
   only with a one-sentence reason why it lowers the HARD max_k overlap.
3. Evals 3-5 -- distinct candidate families.  Each remaining evaluation funds a
   construction that differs in value scheme (sigmoid vs binary/ternary h),
   support structure (block count/widths/positions), or optimizer.
## Switch rule
After any valid evaluation, if the improvement over your best c5 is <= 1e-4
(or the run regressed or errored), do NOT micro-tune that family again.
## Reporting discipline
The training soft-max/logsumexp is a surrogate; the score is the HARD max
recomputed cleanly on the final h after projecting int h = 1 to ~1e-10.
Report the dominant shift k* and the top 2-3 c(k) values before deciding.

## Procedure                                            -- fast-structural-c5-search
1. Rank structures cheaply: block_family_sweeper at n_intervals 300-800 over
   >= 3 settings (n_bumps, spacing in {periodic, clustered, pair}, 2-4 widths).
2. Diagnose the top family: overlap_spectral_diagnostic -> worst shifts k, gap
   to the second-worst, dominant correlation mode.
3. One structural edit: change the shape of h (bump count, width, spacing),
   not the training budget, unless the diagnostic shows the grid binds.
4. Confirm once with evaluate_solution; on regression revert and take the
   next-ranked family.
\end{lstlisting}

\section{Audited reward-routing case study}
\label{app:routing-case-study}

\begin{figure}[H]
\centering
\includegraphics[width=\textwidth]{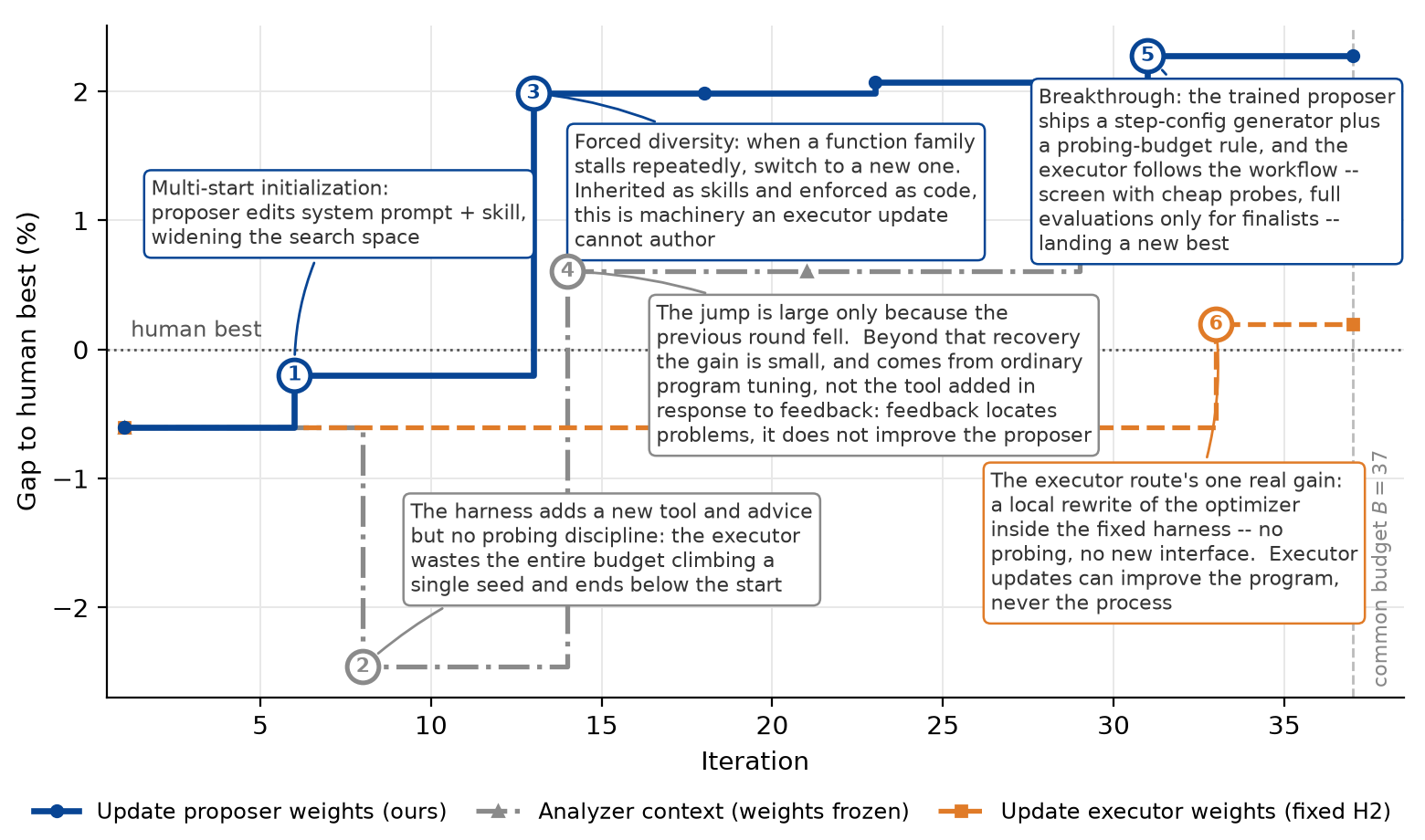}
\caption{\textbf{Autocorrelation~II under three reward routes.}
Numbered markers identify six internally audited events. The internal record
associates each event with harness files, executor trajectories, a program diff,
and content hashes; these records are not part of the current public release.}
\label{fig:case_study}
\end{figure}

According to the unreleased internal audit, event~1 adds multi-start initialization through edits
to both the system prompt and skill. Event~3 inherits a forced-diversity rule
from earlier skills and also instantiates it as middleware. Event~5 combines a
step-configuration generator with a probing-budget rule and a screen-then-verify
workflow. The corresponding unreleased program diffs associate all
three events with changes in the executable search process rather than only
with post-hoc descriptions.

The context route provides a useful counterexample. At event~2, a new tool and
analyst advice are present, but the executor spends its budget on one seed and
finishes below the starting score. Event~4 largely recovers that loss; the
residual gain coincides with re-initialization and parameter-tuning changes
rather than documented use of the suggested tool. The executor-update route's
event~6 gain coincides with a local optimizer rewrite inside the fixed harness.
These observations are consistent with the quantitative ordering in
Figure~\ref{fig:ablation}, but a six-event audit cannot establish that the
identified mechanisms are necessary or sufficient for the score changes.